\documentclass[journal]{IEEEtran}
\usepackage{amsmath,amsfonts}
\usepackage{algorithmic}
\usepackage{algorithm}
\usepackage{array}
\usepackage[caption=false,font=normalsize,labelfont=sf,textfont=sf]{subfig}
\usepackage{textcomp}
\usepackage{stfloats}
\usepackage{url}
\usepackage{verbatim}
\usepackage{graphicx}
\usepackage{cite}
\usepackage{multirow}
\usepackage{svg}
\usepackage[normalem]{ulem}

\usepackage{xspace}
\usepackage{bm}

\makeatletter
\DeclareRobustCommand\onedot{\futurelet\@let@token\@onedot}
\def\@onedot{\ifx\@let@token.\else.\null\fi\xspace}

\def\etc{\emph{etc}\onedot}
\usepackage{array}
\newcolumntype{?}[1]{!{\vrule width #1}}

\usepackage{wrapfig}

\newcommand{\etal}{\textit{et al.}}

\newcommand{\eg}{\textit{e}.\textit{g}., }

\usepackage{array}
\usepackage{booktabs}
\usepackage{mathrsfs}

\usepackage{xcolor}
\DeclareMathAlphabet{\mathcal}{OMS}{cmsy}{m}{n}
\SetMathAlphabet{\mathcal}{bold}{OMS}{cmsy}{b}{n}

\newenvironment{bluepar}
{\color{blue}}
{}

\begin{document}

{\color{red}

}

\newpage

\title{Asynchronous Tracking, Optical Communication and 3D Motion Capture using Event-based Sensors}

\author{Ziwei Wang, Angus Apps, Holly Battisson, Iain Guilliard, Timothy Molloy and Robert Mahony
	\thanks{The authors are with the Systems Theory and Robotics Group,
				College of Engineering and Computer Science,
				Australian National University.
				\tt\{ziwei.wang1, angus.apps, u6070039@alumni, iain.guilliard, timothy.molloy, robert.mahony\}@anu.edu.au}
	}

\markboth{Journal of \LaTeX\ Class Files,~Vol.~14, No.~8, August~2021}
{Shell \MakeLowercase{\textit{et al.}}: A Sample Article Using IEEEtran.cls for IEEE Journals}

\maketitle

\begin{abstract}
Inter-robot communication is a key enabling technology in cooperative robotics applications.
While wireless communication is ubiquitous, simple, and effective, it has inherent limitations: the receiving robot cannot easily localise the source of an incoming signal, clock synchronisation is challenging, and signals are broadcast to all nearby devices rather than targeted recipients.
Optical communication provides a complementary channel that is inherently directional and spatially localised, directly addressing these limitations.
Event cameras are bio-inspired dynamic vision sensors that respond to changes in image intensity with high temporal resolution, high dynamic range, and low latency, making them well-suited as receivers for high-rate optical communication in cooperative robotic systems.
In this paper, we propose the Asynchronous Tracking and Optical Communication (ATOC) system, which integrates LED smart-beacon modulation, event-based detection, optical tracking, and event-data demodulation in a single pipeline.
By simultaneously tracking and demodulating multiple LED smart beacons, ATOC transforms a conventional visual marker into a robust communication channel suitable for a wide range of high-impact robotics applications.
We validate ATOC in a suite of laboratory studies and two `applications': a smart-city demonstration and a 3D motion-capture system.

\end{abstract}


\section{Introduction}

\begin{bluepar}
	\begin{figure*}[t!]
		\centering
		\begin{tabular}{c}
			\\	\includegraphics[width=1\linewidth]{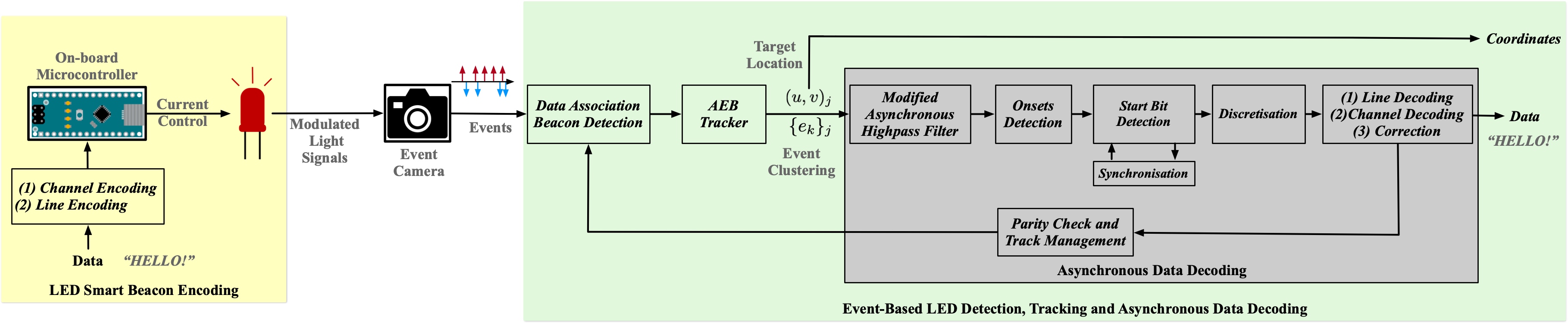}
			\\
		\end{tabular}
		\caption{
			Asynchronous Detection, Tracking and Optical Communication Architecture.}
		\label{fig:pipeline}
	\end{figure*}
\end{bluepar}

Visual tracking of stationary and moving beacons or markers is an enabling technology across a wide range of robotics applications, including warehouse and logistics systems, cooperative automotive leader-follower systems, distributed robotic networks, visual motion capture, \etc.
In most existing systems, beacons are distinguished either by their appearance, such as ArUco markers, colour patterns, or barcodes, or by active (visible or ultraviolet) light, identified through colour, geometric arrangement, or flicker signals.
While such markers can be recognised, they cannot communicate: they convey nothing beyond their own identity.
This is an increasingly important limitation in distributed robotics, such as cooperative driving, smart cities, and formation or swarm control, where agents must continually exchange identity and state information.
Beacons that can be tracked and also transmit data are especially valuable where radio-frequency (RF) communication is unreliable or impractical due to interference, multipath, background noise, bandwidth saturation, intentional signal jamming, or in challenging transmission environments, such as underwater.
Optical Camera Communication (OCC) is a popular alternative for short-range, line-of-sight wireless communication.
Lightweight OCC channels are relatively easy and cheap to integrate into existing robotic systems~\cite{joubert2019characterization}.
Beyond the communication channel itself, OCC also provides a cost- and power-efficient method for localisation, an important consideration in cooperative robotics applications.
Existing OCC systems typically rely on traditional CMOS or CCD cameras to capture the modulated blinking of LED transmitters.
The maximum data rate achievable with standard off-the-shelf frame-based cameras is low, at most 100 to 250 bits per second (bps)~\cite{saeed2019optical}, and most systems report rates well below this.
Higher data rates can be achieved with advanced high-speed cameras or more sophisticated transmitter-receiver setups that allow image sensors to capture down-sampled high-frequency information~\cite{luo2018undersampled}.
However, these systems demand substantial computational resources or dedicated physical infrastructure~\cite{cahyadi2020optical}.
Furthermore, the short exposure times of high-speed cameras require brighter LED transmitters, reducing beacon power efficiency and limiting their effective range.

In contrast to traditional frame-based cameras that accumulate irradiance for all pixels within the exposure time, event cameras continuously monitor the logarithmic light intensity at each pixel independently, generating events only when brightness changes exceed a predefined threshold.
This architecture provides microsecond temporal resolution, high dynamic range (over 120 dB), reduced motion blur, and lower power consumption.
These characteristics make event cameras ideal for integration into existing sensor suites~\cite{wang2021stereo}, capable of detecting high-frequency modulated light from smart beacons for high-speed, low-latency communication between autonomous robots.

In this paper, we propose the Asynchronous Tracking and Optical Communication (ATOC) system, which integrates LED smart beacons emitting a modulated optical signal that is tracked, demodulated, and decoded by asynchronous event processing.
The high temporal resolution and sensitivity of event cameras enables high baud rate communication (relative to existing OCC systems) along with accurate high-speed tracking of the visual beacon.
We exploit the flickering nature of the beacon to detect candidate beacons as event-blobs in the data stream ~\cite{wang22icra}, and then track beacons using the AEB (Asynchronous Event Blob) tracker~\cite{wang2024aeb}.
Events associated with each tracked beacon are processed using a modified event-based high-pass filter~\cite{scheerlinck2018continuous} to reconstruct the transmitted intensity signal asynchronously.
An onset detection algorithm then extracts the binary bitstream, automatically synchronises with the start delimiter, and decodes the transmitted message.
We transmit ASCII-encoded messages using Modified Frequency Modulation (MFM) line coding, with an extended Hamming code for error correction and a dedicated start delimiter for framing and synchronisation.
The communication pipeline is designed to be simple, robust and computationally efficient.

We validate ATOC through a series of indoor and outdoor experiments that encompass both controlled performance evaluation and real-world robotic applications.
A spinning-disk experiment characterises communication accuracy under varying transmission rates and image-plane velocities, demonstrating error-free transmission at 12.5~kb/s for motion up to 3.5~kpx/s and at 11.11~kb/s for motion up to 9.2~kpx/s.
The system also detects and validates newly visible beacons within 5~ms.
In an outdoor bicycle experiment, ATOC decodes 138 ASCII characters during a single 327~ms observation, demonstrating its potential for smart-city applications.
Finally, we use ATOC to achieve 3D visual motion capture and demonstrate this by capturing a musician playing a quebecois fiddle tune.
In the two-camera setup, a packet counter in the transmitted message allows global temporal synchronisation of independent event cameras with a precision of 4.54~$\mu$s, enabling 3D reconstruction with a precision of 2~mm at approximately 4~kHz.
The resulting track quality matches or exceeds state-of-the-art motion capture systems at a fraction of the complexity, requiring no central synchronisation hardware and no cabling between cameras.
The total hardware cost, including event cameras and inexpensive LED beacons, makes this potentially one of the simplest and highest-performance visual motion capture systems possible with current technology.

In summary:
\begin{itemize}

	\item We propose the Asynchronous Tracking and Optical Communication (ATOC) system that integrates visual beacon detection and tracking with broadcast optical camera communication, and demonstrate its capability for high-rate data communication in real-world experiments.

	\item We develop a robust event-based optical communications pipeline, including message encoding, beacon detection and tracking, signal onset and start-delimiter detection, synchronisation, and bit error correction.

	 \item We validate ATOC through quantitative characterisation across a range of transmission rates and image-plane speeds, and through two real-world applications: a smart-city deployment and a wireless 3D motion-capture system in which a packet counter broadcast by each beacon synchronises multiple uncabled event cameras for high-rate 3D reconstruction.

\end{itemize}

The beacon PCB designs and all code developed will be provided as open source resources when the paper is published.

\section{Related Work}
There is an established literature in Optical Camera Communication (OCC) exploiting LED light modulation~\cite{saeed2019optical}.
Different keying technologies investigated include under-sampled frequency shift on-off keying (UFSOOK)~\cite{roberts2013undersampled}, camera multiple frequency shift keying (CM-FSK)~\cite{rajagopal2014visual,lee2015rollinglight}, pulse width modulation/pulse position modulation (PWM/PPM)~\cite{aoyama2015line,aoyama2015visible}, spatial-2-phase-shift keying (S2-PSK)~\cite{nguyen2016region}, camera on-off keying (C-OOK)~\cite{nguyen2016high}, \etc.
There are also several established standards (\eg IEEE 802.15.7) for short-range optical wireless communications~\cite{saeed2019optical}.
Usually, to capture the high-frequency LED blinking signal, a high-speed camera is required, up to 1000 fps~\cite{roberts2014automotive,ebihara2015layered}.
Lower frame rate CMOS cameras with the selective capturing strategy~\cite{teli2018high}, such as under-sampled modulation~\cite{luo2014undersampled} or other custom-built systems~\cite{oike2003smart,imai2016performance} have also been investigated.
LiFi technology offers much higher speed and lower latency, but requires strict alignment of photodiodes, dedicated hardware, and faces challenges including poor robustness to motion, limited transmission distance (typically 1--10~m indoors), and difficulty in handling multiple sources.

In the robotics field, researchers have recognised the potential of event cameras to create low-cost optical communication channels for embedded and robotic systems.
Wang~\etal~\cite{wang2022smart} demonstrated effective, lossless transmission at 500~bps in full sunlight at a distance of 100~m.
Subsequent research has introduced improved algorithms and processing pipelines, leading to enhanced tracking and communication performance~\cite{aranda2024enhancing,von2024dynamic}.
Recent work has also explored parallel transmission channels using digital micro-mirror devices (DMDs)~\cite{wang2024towards} and 2D dynamic markers~\cite{su2024motion, kitade2024bicode} to further increase transmission rates.
Joubert~\etal~\cite{joubert2019characterization} characterised an event camera LED setup for event-based imagers applied to modulated light signal detection.
Some preliminary results were also obtained by Perez-Ramirez~\etal~\cite{perez2019optical} where events were first accumulated into synchronous frames.
They then proposed a start frame delimiter (SFD) and symbol design derived from the Composite Waveform, and showed some Bit Error Rate (BER) results for the transmission of 2-byte packets.
Soga~\etal~\cite{soga2025evaluation} proposed an algorithm for OCC using LED bars, where clusters are uniquely identified using Walsh-Hadamard code pilot sequences.
Tofighi~\etal~\cite{tofighi2026fast} assigned different pulse patterns to codewords in a static look-up table.
Although the number of pulses is assigned such that the most common codewords use fewer pulses, the transmission rate remains variable depending on the message.

A key requirement for embedded and robotic system optical communications is the ability to track the source of transmission.
General event-based tracking algorithms have an extensive literature including pose estimation and close-range tracking~\cite{faessler2014monocular}, with applications in multi-agent unmanned aerial vehicles~\cite{dias2016board,walter2018mutual,walter2018fast} and autonomous underwater vehicle systems~\cite{bosch2016close,gracias2015pose}.
Tracking fast blinking LED markers is a natural specialisation of the general algorithms.
Censi~\etal~\cite{censi2013low} proposed a pose tracking algorithm using four infrared LED markers with different blinking frequencies which are distinguishable by event cameras.
They accumulated triggered events in short time windows and tracked the pose of a drone fitted with LED markers using a particle filter.
Instead of the traditional CMOS cameras, Chen~\etal~\cite{chen2020novel} used an event camera in their visible light positioning system with multi-LEDs, which largely reduced the computational cost and latency.
They estimated different frequencies of LED markers based on the time intervals between positive-negative transitions at a pixel and then tracked LED markers with their GM-PHD filter, resulting in an accuracy of 3~cm when the distance between LEDs and camera is less than 1~m.
More recently, monocular systems have employed LED beacons flashing at distinct frequencies, mounted on agile quadrotors~\cite{bauersfeld2025monocular} or on rigid boards~\cite{ebmer2024real}, to enable real-time, PnP-based motion capture.

Combining tracking and optical communications is a natural next step for embedded and robotic systems.
Soga~\etal~\cite{soga2026real} extended their OCC work~\cite{soga2025evaluation} to perform real-time tracking and communication from multiple LED bars.
Our early work on smart visual beacon systems~\cite{wang2022smart} already had a tracking module that has been extended to manage asynchronous detection, tracking, and track management of small blob-like objects, enabling high-speed, real-time tracking of LED beacons~\cite{wang2024aeb, apps2025aemot}.
The present paper extends our preliminary work~\cite{wang2022smart,apps2025aemot} to improve the communications protocols and exploit the multi-target detection and tracking capability.

\section{Asynchronous Tracking and Optical Communication (ATOC) Architecture}
Our ATOC system (Figure~\ref{fig:pipeline}) consists of three main steps: (1) LED smart beacon encoding (yellow); (2) event-based detection and tracking (green); (3) asynchronous data decoding (grey).
The encoding and decoding are done on separate systems, \eg encoding is undertaken on the beacon embedded compute and decoding is on the mobile robot central processor or a dedicated sensor compute system if it is available.

In robotics applications, it is often the location of the beacon that is most relevant to the application, and the track quality is as important as the data transmitted.
That said, the communication channel provides key information such as beacon ID, state or timing information, or other short messages.
As such, the communication protocol is designed for single direction (simplex) transmission of a relatively short message, of the order of 100 bytes of data or less.
Each message is divided up into ``Hamming packets'' of three bytes of data (as discussed below) and encoded using the extended Hamming (32,26) code.
The packet size is a trade-off between data rate (larger packets) and dropout (visibility of the beacon is only guaranteed for short periods in many robotics applications).

\subsection{LED Smart Beacon Encoding}
\label{sec:Encoding}

\begin{figure}[ht]
		\centering
		\includegraphics[width = \linewidth]{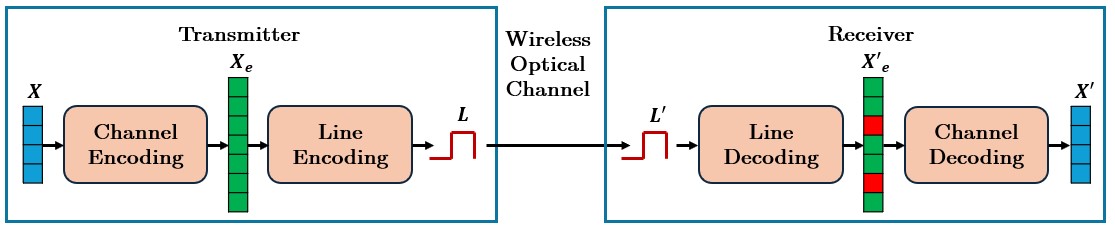}
	\vspace{-1em}
	\caption{Communications pipeline.
		In the transmitter, a message is divided into data packets $X$ that are first encoded with parity bits via channel coding, producing the encoded packet $X_e$. This is converted into LED high-low signals $L$ through line coding and transmitted.
		In the receiver, an event camera captures the optical signals and a detection and tracking algorithm is used for data association to discard background visual cues and recover the binary stream $L'$ of the observed LED.
		Line decoding reconstructs the received encoded packet $X_e'$, followed by channel decoding to correct errors and recover the data packet $X'$.
		These data packets are then reconstructed into the received message.
		}
	\label{fig:CommunicationsChannel}
\end{figure}

A message is divided up into packets $X \in \{0,1\}^\ell$ of data, each of which will be encoded and sent over the optical communications channel (Figure~\ref{fig:CommunicationsChannel}).
The data packet $X$ is encoded to obtain an encoded packet $X_e \in \{0,1\}^n$ with an additional $n-\ell$ parity bits used for error detection and correction.
The line coding protocol translates the encoded packet $X_e$ into a signal $L$ transmitted by the LED.
The event camera observing the LED captures a signal $L'$ as a flickering spatio-temporal blob~\cite{wang2024aeb} in the event stream.
The detection and tracking algorithms undertake data association that identifies events associated with the signal $L'$ and discards those associated with the background image.
The inlier events are used to reconstruct a signal and the onset points of the original LED transitions are identified.
The resulting time-sequence is used to extract an estimate of the received encoded packet values $X_e' \in \{0,1\}^n$.
Channel decoding is used to identify and where possible correct bit errors, and to discard the parity bits, in order to reconstruct a received data packet $X' \in \{0,1\}^\ell$.
If error correction has been successful, the received packet $X'$ should be identical to the transmitted data $X' = X$.
Finally, the full message is reconstructed from a sequence of packets.

\subsubsection{MFM Line Encoding}\label{sec:Line_encoding}
Line encoding defines how digital data is physically represented in the transmitted signal.
We use the Modified Frequency Modulation (MFM) line encoding protocol, a widely used run-length limited (RLL) line coding scheme~\cite{immink1990runlength}.
In our system, the signal is generated by an STM32 microcontroller (see hardware details in Appendix~\ref{appendix: hardware}), which modulates the voltage to control the rapid blinking of the LED beacons.
Under MFM the bit rate is half the modulation (or switching) frequency of the LED\footnote{
To avoid confusion with the data bits we will describe the LED's ON state as a high voltage level (`H') while the OFF state corresponds to a low voltage level (`L').
The data bits themselves will be referred to as logical bits `0' and `1'.
}.
A `1' bit is always encoded as a transition from low to high (\{L H\}).
A `0' bit is encoded as \{L L\} if the preceding bit is `1', and as \{H L\} if the preceding bit is `0'.
The encoding pattern is summarised in Table~\ref{tab:mfm_encoding}.
This encoding maps binary data onto the LED's physical states, enabling the transmission of information through its ON/OFF patterns.
MFM coding guarantees no more than three consecutive OFF (low-voltage) LED states, ensuring the beacon appears as a constant light source to human eyes, while enabling robust detection and tracking in the event stream.

\begin{figure}[!ht]
		\centering
		\includegraphics[width=\linewidth]{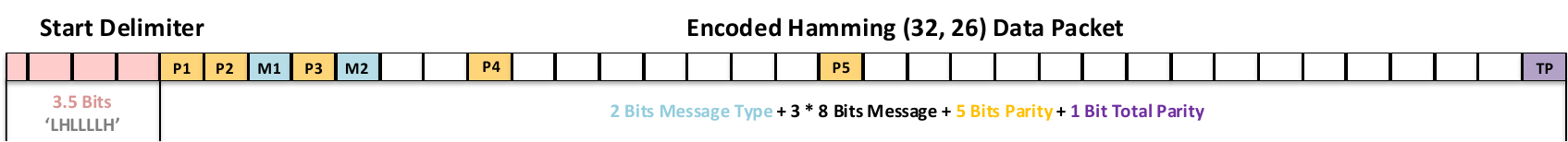}
	\vspace{-1em}
	\caption{
		Encoded communication packet using extended Hamming (32,26).
		Pink bits are associated with the start delimiter \{LHLLLLH\} and are not part of the Hamming codeword.
		Two bits (blue) are used as a message type tag and then 24 bits (3 bytes) (white) contain the primary data packet.
		Five parity bits (orange) and one overall parity bit (purple) are distributed throughout the message following the Hamming code structure \cite{Hamming_1950}.
		}
	\label{fig:EncodedPacket}
\end{figure}

Each encoded packet is preceded by a 3.5-bit start delimiter pattern $\{\text{LHLLLLH}\}$ (see Figure~\ref{fig:EncodedPacket}) that marks the beginning of a packet.
The start delimiter pattern is not a valid bit stream and will never occur in the data stream, providing a unique identifier to allow synchronisation of the decoder.
The start delimiter detection subroutine searches for this pattern and synchronises the bit stream on the \{LLLLH\} transition (that also has excellent time resolution due to the long OFF sequence that allows clean detection of the HIGH transition).

\begin{table}[h]
	\centering
	\caption{Modified Frequency Modulation encoding scheme. A logical `1' is always encoded as `LH'. A logical `0' is encoded as `LL' if it follows a `1', or `HL' if it follows a `0'.}
	\begin{tabular}{|c|c|c|}
		\hline
		\textbf{Data Bit} & \textbf{Previous Bit} &\textbf{Encoded LED States} \\
		\hline
		1 & x & L H \\
		\hline
		0 & 1 & L L \\
		\hline
		0 & 0 & H L \\
		\hline
	\end{tabular}
	\label{tab:mfm_encoding}
\end{table}

\subsubsection{Channel Encoding}
\label{sec:channel_encoding}

Channel encoding is the mechanism used to convert the message data into a serial communication packet, shown on the left of the transmitter block of Figure~\ref{fig:CommunicationsChannel}.
We employ the extended Hamming code (32, 26) for single-error correction and double-error detection~\cite{Hamming_1950}.
Each encoded packet is 32 bits, comprising 26 data bits and 6 parity bits.
The default configuration for our protocol is to allocate 2 bits for a message type field and then encode three bytes of data in the remaining 24 bits.
For dedicated communications channels, the message type bits are useful for splitting larger messages into multiple packets that can be reconstructed asynchronously as we show in the smart city demonstration \S\ref{sec:smart_city}.
It can also be used to specify types of data, for example, ID, velocity, position, \etc, for applications with real-time feedback requirements.

Six parity bits improve transmission reliability by providing forward error detection and correction.
Each parity bit is computed based on the bit positions using even parity.
This coding structure allows the receiver to detect and correct single-bit errors, and identify uncorrectable errors (when more than 1-bit error occurs in a packet).
Beyond error correction, the parity bits also play a key role in packet detection and decoding by providing feedback to the detection algorithm, distinguishing beacon signals from background noise, and aiding in synchronisation (see Figure~\ref{fig:pipeline} and Section~\ref{sec:Detection algorithm}).

The full transmitted packet is length 35.5 bits containing 26-bits of data leading to a block code rate of 73.2\%.
That is, the start delimiter and error correction capabilities consume 26.8\% of the communication bandwidth.
The block code rate could be increased by using a longer Hamming code since the number of data bits scales faster than the required parity bits.
However, longer codes require longer observation of the data signal
compromising their use in robotics applications where the beacons may only be observed for short periods of time.
The proposed protocol transmits a packet in roughly 3.5~ms at $\sim$10~kb/s (cf.~Tables \ref{tab:packet_accuracy_static} and \ref{tab:packet_accuracy_spin}).
Shorter Hamming codes could also be used, and indeed we originally trialled a (16,11) Hamming code that is transmitted in only 2~ms.
However, the block code rate for a (16,11) Hamming code is only 56\%, significantly impacting the data throughput.
We propose the (32,26) Hamming protocol as a good compromise for embedded and dynamic robotic systems applications.

\subsection{Event-Based LED Detection, Tracking and Asynchronous Data Decoding}
\label{sec:LED_detection_tracking}

We use an event camera to track and capture the modulated LED blinking signal.
We adopt the track management pipeline of AEMOT (Asynchronous Event-based Multi-Object Tracking)~\cite{apps2025aemot} to monitor the status of each target in which each individual target is followed by an AEB (Asynchronous Event Blob) tracker~\cite{wang2024aeb}.
For detection, we threshold the Surface of Active Events (SAE)~\cite{Benosman14tnnls}.
The pipeline generates a stream of events associated with a given target, and this target-specific event stream is post-processed to generate a high-pass intensity signal on which leading edge detection is undertaken.
The resulting timestamps of leading edge detections are converted to the observed ON-OFF LED pattern and decoded.
The AEB tracker also provides the location and optical velocity of the observed target.

Although event cameras have high temporal resolution and low latency, they are still subject to refractory period limitations and nonlinear effects~\cite{Wang19acra,wang2021asynchronous}.
On most existing cameras, events are also collected and processed through an FPGA and this introduces temporal quantisation and jitter.
For common cameras, the timestamps are typically provided at microsecond resolution and are accurate to 10--100~$\mu$s depending primarily on event rate and ambient light but also subject to other environmental factors.
These limitations induce limits on the communication bandwidth we discuss in the experimental section (see Table~\ref{tab:packet_accuracy_static} and \ref{tab:packet_accuracy_spin}).

\subsubsection{Detection and Tracking of Beacons}\label{sec:Detection algorithm}
An event camera observes events across the full image.
Within the camera field of view there may be multiple beacons visible, each appearing as a localised flickering blob, as well as background events.
The background events may also contain flickering blobs that are not beacons, for example, LED lights on cars or other devices flicker at high rates and generate beacon like visual signals in the event camera.
The proposed algorithm detects potential blobs, tracks each blob, and then validates the track based on decoding meaningful packets of data from its event stream.

Each incoming event is checked against existing tracks in the tracker, firstly against tracks that are known to be beacons, and secondly against other tracks that are potential beacons or known non-beacons that are being tracked to remove their events from the detection process.
Only events that are not associated with existing tracks are passed to the detection algorithm.
Initial detection of possible target beacons is based on comparison to the Surface of Active Events (SAE)~\cite{Benosman14tnnls}
containing the timestamp of the last event that occurred at each pixel.
As the expected event rate is high, particularly with multiple beacons, the SAE-based detection is designed to be computationally lightweight.
A time threshold parameter and minimum blob size parameter are used and must be tuned for the application; a scenario with few potential targets that may be distant from the camera may accept single-pixel blobs, while a scenario with many flickering sources, in which only those close to the camera are relevant, would only accept multi-pixel blobs.
Each incoming event is checked against the existing SAE values of firstly its own pixel, and then against neighbouring pixels.
If there is a contiguous blob of pixels including the event pixel
in the SAE that all have timestamps within the time threshold, then the event triggers the detection algorithm; otherwise, the SAE is updated and the next event is considered.
The detection algorithm ensures that only blobs of minimum blob size are detected.
Note that all future events that are close (within twice the estimated variance of the event blob) to this detected blob will now be associated with the candidate track and not trigger the detection algorithm.

Once a detection is confirmed an instance of the AEB tracker is initialised to track the flickering blob.
The AEB tracker~\cite{wang2024aeb} is a lightweight Kalman filter that estimates the size and shape of the event blob as well as its position and velocity.
These parameters are used to associate incoming events to the given track allowing the tracker to continue to adapt to motion and shape changes of the event blob.
All events associated with a given track are first processed in the AEB tracker and then passed as an event stream to a signal reconstruction block
(cf.~\S\ref{sec:Asynchronous high-pass filter and threshold}).
The resulting signal is processed  (cf.~\S\ref{sec:PeakDetection})  and decoded (cf.~\S\ref{sec:Discretisation and decoding}) to extract data.
The track validation criterion is based on extracting a valid parity check for packets received; that is, when a full packet is received with valid parity bits then the track is upgraded to a communication beacon and that track is prioritised for data association.
A track that does not produce a valid parity check in a specified time period is labelled as non-valid.

Depending on computational load, non-valid tracks may be maintained in the AEB tracker without continuing to reconstruct the underlying intensity signal.
This has the advantage that associated events are not considered new detections, as well as providing contextual information to a robotic system about moving objects such as LED lights, \etc, in the environment.
Alternatively, non-valid tracks are terminated.
If the track was associated with a flickering source in the image, then the detection algorithm will immediately pick up a new detection, initialise a new track in the AEB, and lead to a new instance of validation, eventually terminating the track again.
The process takes a finite time and the sequence repeats ad infinitum.
As long as sufficient computational resources are available, repeatedly redetecting non-valid event blobs and checking them for validity ensures that no new communication beacon is missed due to legacy tracks.
If non-valid tracks are monitored in the AEB, then it is wise to include a routine that regularly revisits non-valid tracks and checks for a valid packet transmission to ensure that no valid beacons are ignored due to previous misidentification.

\subsubsection{Signal Reconstruction}
\label{sec:Asynchronous high-pass filter and threshold}

The event stream is integrated through a high-pass filter to reconstruct a copy $L'$ of the original signal.
In this paper, we use a modified asynchronous event-based high-pass filter developed in~\cite{scheerlinck2018continuous}, which attenuates low frequency drift in the data stream while preserving high frequency data variations (cf.~Figure~\ref{fig:static_10kbps_16kbps}).
Consider a sequence of events $e_k := (\xi_k, t_k, \sigma_k) $ for $k \geq 0$, where $\xi_k = (u_k,v_k)$ denotes pixel location, $t_k$ the timestamp, and $\sigma_k \in \{-1, +1\}$ the event polarity, associated with a tracked smart beacon.
This high-pass filter is implemented asynchronously, with updates computed only upon event arrival.
The intensity $L'$ of a beacon is updated for each event $e_k$ that is associated with that beacon
\begin{align}
L' (t_k) =\begin{cases}
\exp(-\overline{\alpha} (t_k - t_{k-1})) L'(t_{k-1}) + \sigma_k , &\ \sigma_k = +1, \\
\exp(-\underline{\alpha} (t_k - t_{k-1})) L'(t_{k-1}) + \sigma_k , &\ \sigma_k = -1, \\
\end{cases}
\end{align}
where $\overline{\alpha}$ and $\underline{\alpha}$ are gains that govern the filter high-pass roll-on frequencies for rising and falling signals, respectively.
Because event cameras usually have a positive bias that triggers more positive events than negative ones due to the undesirable junction leakage in the amplifier circuit or other noise~\cite{Wang19acra}, the detection of rising edges is more robust than that of falling edges.
As a consequence, we focus only on detecting the rising edge of LOW-HIGH transitions in the signal and use these to reconstruct the data as discussed in \S\ref{sec:Discretisation and decoding}.
To improve sensitivity to the rising signal, we use a significantly larger decay rate for the negative events than for the positive events  $\underline{\alpha} \gg \overline{\alpha}$.
This has the effect of driving the signal back to zero quickly during a sequence of negative events and resetting the signal for the next rising leading edge.

\subsubsection{Leading Edge Detection}
\label{sec:PeakDetection}
\begin{figure}
	\centering
	\subfloat{\label{subfig:a} \includegraphics[width=0.48\columnwidth]{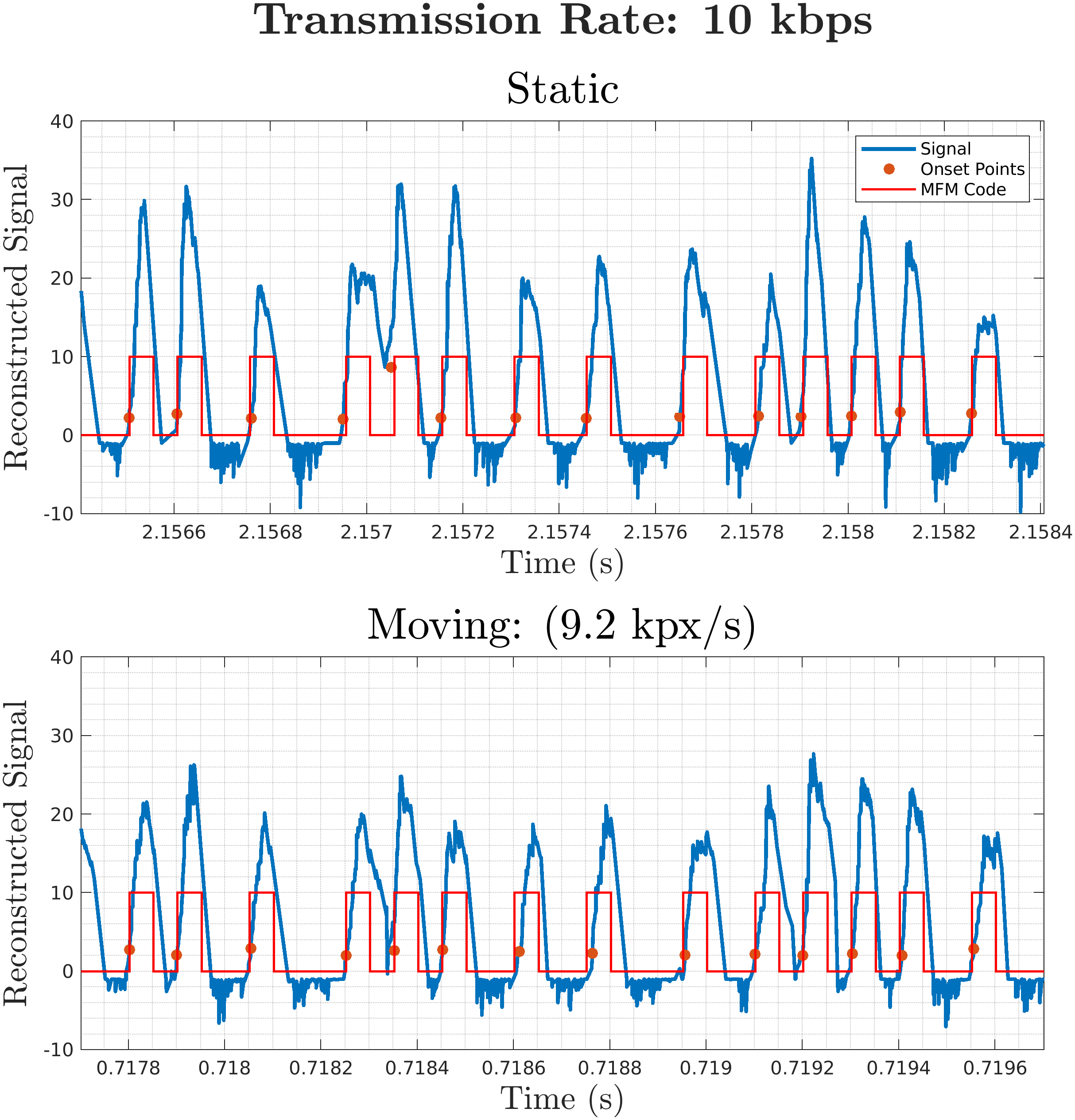}}
	\subfloat{\label{subfig:b} \includegraphics[width=0.48\columnwidth]{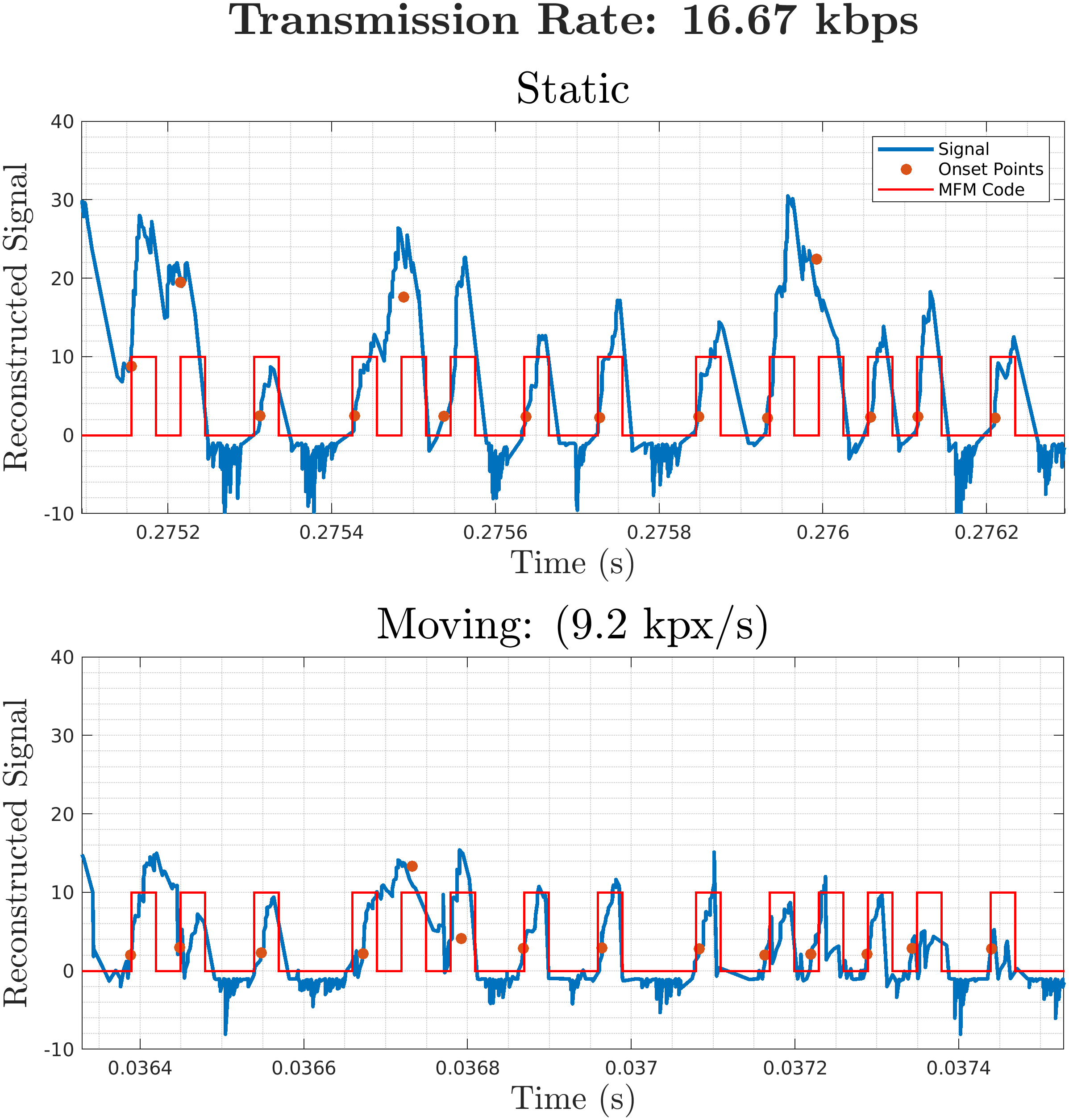}}
	\vspace{-3mm}
	\caption{
	Example of modified high-pass filter intensity signals from a smart beacon at different transmission rates (10~kb/s and 16.67~kb/s), shown in both static and moving states (9.2~kpx/s).
	Detected onset points (dots) and the ground truth MFM (square wave) are overlaid as red.
	Note that at 16.67~kb/s the data transmission is highly compromised with only $\sim$60--65\% bit accuracy and only $\sim$40--45\% message accuracy when static (cf.~Tables~\ref{tab:packet_accuracy_static} and \ref{tab:packet_accuracy_spin}).
	Conversely at 10~kb/s transmission the bit and message accuracy are 100\% even when the LED is moving in the image.
	}
	\label{fig:static_10kbps_16kbps}
\end{figure}

As event cameras respond to relative changes in light intensity, peaks in the reconstructed signal correspond to periods when the LED is in the HIGH state.
Due to the very high $\underline{\alpha}$ gain, the signal during the LOW state of the LED quickly converges to zero (cf.~Figure~\ref{fig:static_10kbps_16kbps}).
Identifying the onset of each peak enables detection of the corresponding LED LOW-HIGH transitions and the time sequence of rising leading edge information can be used to decode the transmitted signal.

As shown in Figure~\ref{fig:static_10kbps_16kbps}, the peaks seen in the raw reconstructed signal $L'$ derived from event data can vary in both maximum intensity and temporal width.
Additionally, intermittent negative events occurring during a positive spike or sparse positive events may distort the signal, resulting in fluctuating rather than sharp peaks.
Nevertheless, a consistent characteristic across all peaks is a sharp positive gradient at the leading edge of a LOW-HIGH transition.
Leading edges are detected by estimating the intensity gradient $m_k$ over a sliding window of fixed length in events,
\begin{equation}
m_k = \frac{L'(t_{k}) - L'(t_{k-n})} {t_{k} - t_{k-n}}
\end{equation}
where $t_k$ is the event time of the most recent element in the sliding window and $n$ is the length of the sliding window.
Note that since the window length is measured in events, rather than in time, it is very sensitive to a rush of events associated with a leading edge and insensitive to a sequence of positive events that occur slowly.
In essence, the criterion is equivalent to searching for a sequence of $n$ positive events occurring in a short time period.
The leading edge time allocated to a detection is the oldest timestamp in the window, $t_{k-n}$, corresponding to when the leading edge started to rise.

Since the transmission rate corresponds to the symbol rate $f_s$, peaks should not occur in rapid succession.
Therefore, once an onset is detected, a refractory period is enforced during which no additional onsets can be recorded.
At high transmission rates, we observe that negative events may not be triggered during a `HIGH-LOW-HIGH' sequence, resulting in a gradual intensity decay driven only by the positive event gain.
This lack of a sharp negative drop can shift the detected onset away from the true transition, leading to decoding errors.
To address this, we designate onset points with a preceding intensity drop below zero as `trusted onsets', and use their timestamps as reference positions.
Non-trusted onsets are then corrected based on the timing of the most recent trusted onset and the known transmission rate.
The reference is updated whenever a new trusted onset is identified.
The gains are chosen such that any LOW-LOW sequence provides enough time for the positive signal to decay far enough for a `trusted onset' to be detected.
Figure~\ref{fig:static_10kbps_16kbps} illustrates the alignment between detected onset points in the reconstructed signal and the actual LED `ON' transitions for a valid encoded packet $X_e$.

\subsubsection{Start Delimiter Detection and Synchronisation}
\label{sec:start_delimiter_detection}

To detect the start of each packet, we use a synchronisation method based on the delimiter pattern \{L H L L L L H\}.
There is no valid encoding pattern that can generate a sequence of four LOW signals in a row; the sequence ``101'' \{L H L L L H\} has only three LOW signals.
The period of a cycle $\frac{1}{f_s}$ (where $f_s$ is the symbol rate) is known and the two leading edges of the HIGH signals in the start delimiter should be $\frac{2.5}{f_s}$s apart providing a clear marker for the start delimiter.

Synchronisation is derived from the second HIGH pulse of the start delimiter.
This leading edge provides highly reliable detection due to the preceding extended LOW state.
Once a start delimiter is identified, a first-in-first-out (FIFO) buffer stores the leading edge timestamps over the 64 LED state slots that follow the synchronisation signal, corresponding to the 32 bits of the transmitted message.
The full timestamp of the leading edge detections is not stored, rather the detections are quantised to the closest timestamp $\frac{1}{2f_s}$, $\frac{2}{2f_s}$, \ldots, $\frac{64}{2f_s}$ corresponding to the onset times as measured from the reliable synchronisation leading edge.
This saves data in the FIFO buffer, since only an 8-bit timestamp needs to be stored, and significantly improves the robustness of the symbol decoding.
Unreliable leading edge detection is typically caused by a repeated LOW-HIGH-LOW-HIGH sequence; because every detection is referenced to the start delimiter, timing errors within such a sequence do not cascade through the remainder of the packet.

Once the buffer is full, the algorithm searches for the next start delimiter.
If the channel is synchronised and working, the next start delimiter should be received immediately.
If not, detection keeps running until the next start delimiter is found.
In high noise environments, it is possible that a whole packet may be missed due to a missed start delimiter, typically due to missing its first LOW-HIGH transition.
Even if this leading edge is missed, the second LOW-HIGH transition is almost always detected (at least if communication is possible at all) and it would be possible to write additional routines to detect such cases.
At the moment, we have only implemented the full start delimiter detection with no additional logic.

\subsubsection{Packet Decoding, Evaluation and Correction}
\label{sec:Discretisation and decoding}

Once complete, the packet stored in the FIFO buffer is read out to the decoding module and the sequence of leading edge detections is used to reconstruct a 64-slot sequence of LED states $\{L H \cdots L H\}$.
Since the timing sequence was quantised when stored, this is a straightforward process.
The MFM decoding scheme is run to obtain a received packet $X_e' \in \{0,1\}^{32}$.
Once $X_e'$ is obtained, \textit{channel decoding} is applied to recover the original message data $X'$.
The extended Hamming code (32, 26) supports single-bit error correction and double-bit error detection.
To determine if a 32-bit sequence is a valid packet, we evaluate the parity bits based on the known packet structure (see Figure~\ref{fig:EncodedPacket}).
A packet is considered valid under two conditions:
(1) all parity checks pass, indicating no error; or
(2) the parity checks identify a single-bit error, which is corrected as described below.
If the parity checks indicate more than one bit error, the correction capability is exceeded.
In such cases, the packet is discarded and an error flag is asserted to the track management system.
The track management system then determines whether to terminate the beacon track or redesignate it as a non-beacon if it consistently produces erroneous packets.

When a single-bit error is detected, we apply a low-complexity \textit{bit-wise XOR correction} method~\cite{Saiz_2016}.
This involves identifying the positions of all bits with value `1' in the 32-bit received packet.
A bitwise XOR operation is performed over these indices, and the result indicates the position of the erroneous bit.
The bit at this index is then flipped to produce the corrected packet $X'$.
This method leverages even parity and is computationally efficient, making it suitable for real-time applications of the proposed communication protocol~\cite{Saiz_2016}.
Finally, the corrected packet is passed to the message module in the firmware of the robotic system.

\section{Experimental Study}
\label{sec:Experimental_study}

In this section we document a set of experimental studies undertaken in lab conditions to quantify the performance of the proposed scheme.
Later, in \S\ref{sec:smart_city} and \S\ref{sec:3D_optical_tracking_and_recon}, we present the real-world ``smart city'' and ``visual motion capture'' case studies.
The first laboratory experiment characterises the communication bandwidth of the architecture and its dependence on motion of the optical beacon in the image.
The second experiment considers the question of detection and tracking of a beacon and demonstrates the speed with which a beacon can be detected, tracked, and the communication decoded.

\begin{figure}[t!]
	\centering
	\begin{tabular}{c c}
		\\	\includegraphics[width=0.45\linewidth]{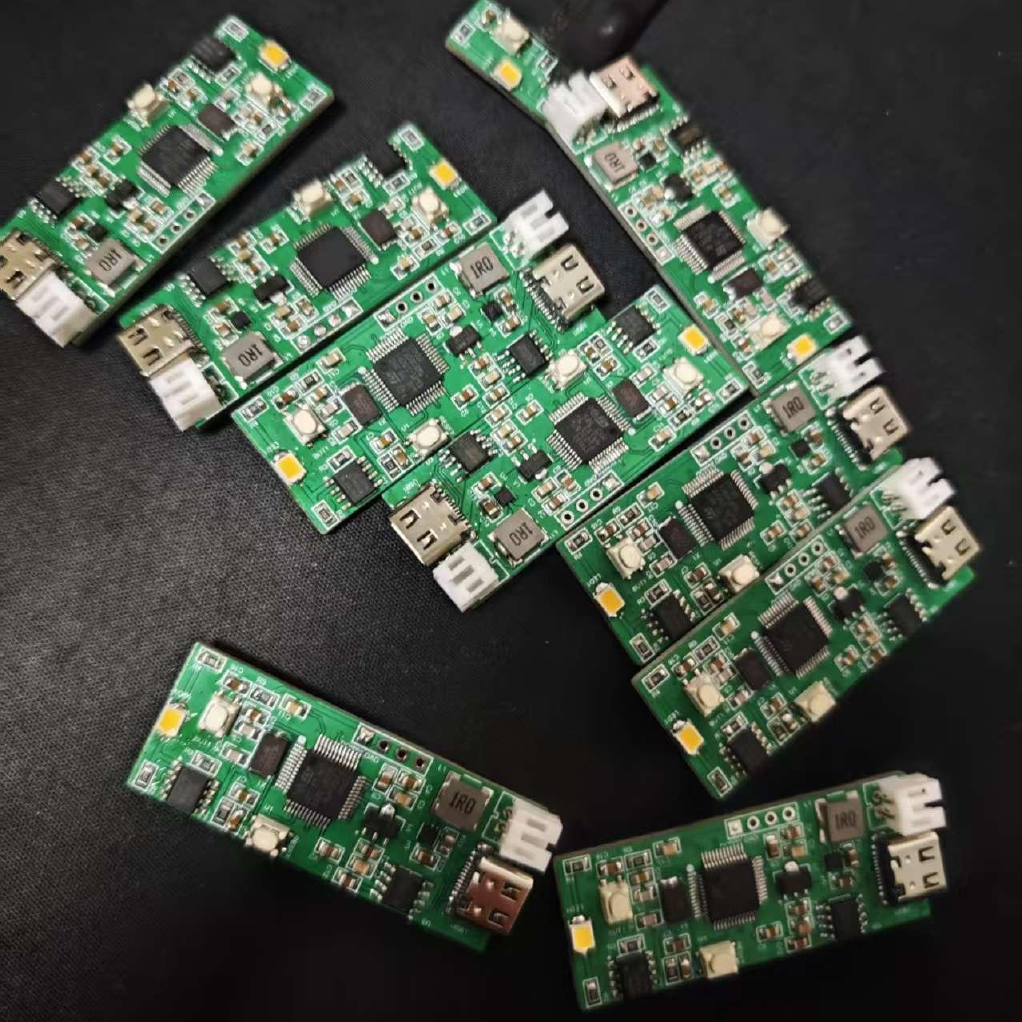} &
		\includegraphics[width=0.45\linewidth]{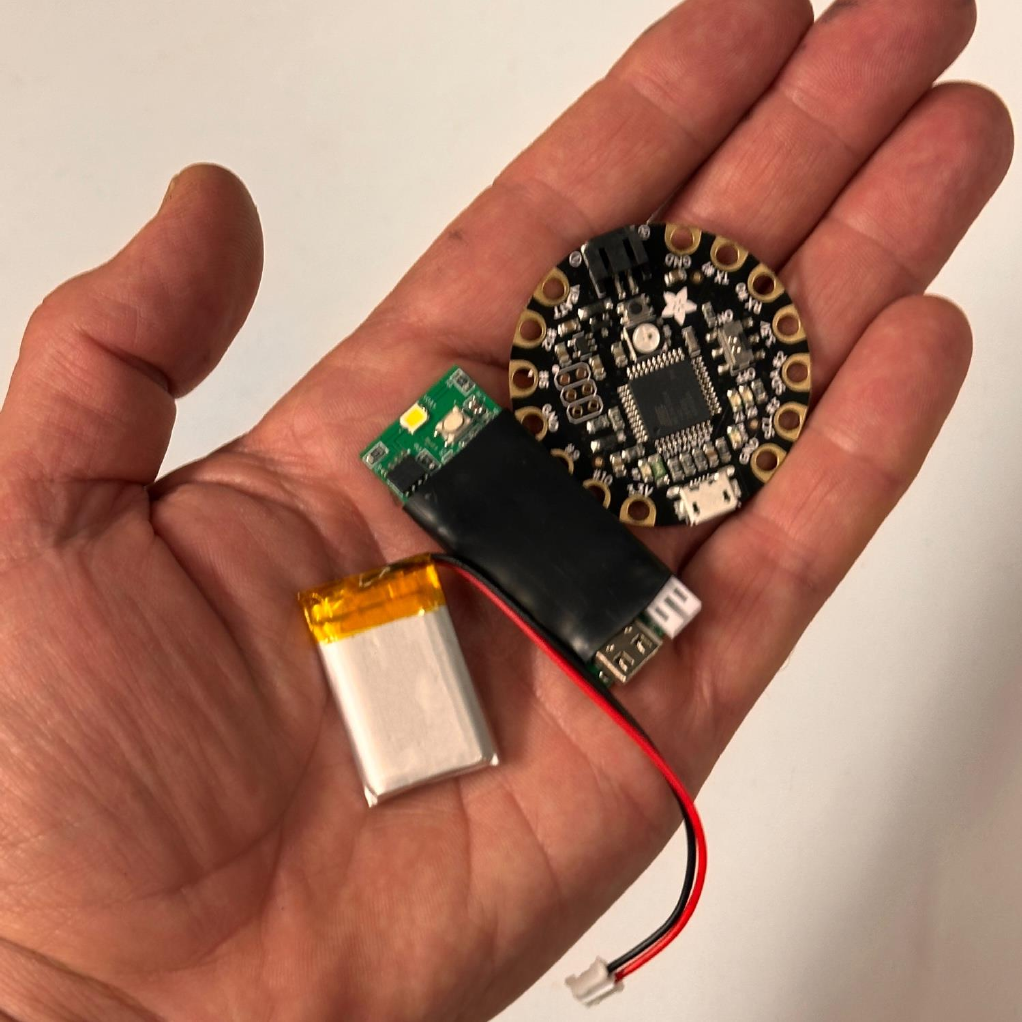} \\
		(a) Our smart beacons  & (b) Adafruit FLORA (right)
	\end{tabular}
	\caption{\label{fig:circuit and camera}
		Demonstration of our designed smart beacons, controlled by an STM32F103 microcontroller.
		We also show the size comparison to the off-the-shelf smart beacon Adafruit FLORA in (b).
		It is controlled by an ATmega32U4, which is only suitable for lower frequency optical communication applications.
	}
\end{figure}

For the experimental programme we designed and built a custom PCB (Figure~\ref{fig:circuit and camera}(a)) with an STM32 microcontroller and a surface-mounted 2835 natural white LED (2.8~$\times$~3.5~$\times$~0.8~mm, 150~mA, 0.5~W) to provide a programmable high intensity light source.
Details of the PCB design are provided in Appendix~\ref{appendix: hardware} and will be released open source with the publication of the paper.
We also evaluated an off-the-shelf commercial smart beacon, the Adafruit FLORA
\cite{2025_AdaFruit} (Figure~\ref{fig:circuit and camera}(b)).
The Adafruit FLORA can only manage low rate communications due to the processor (an ATmega32U4) and lower power LED.
Nevertheless, it is a low cost entry level smart beacon (US\$14.95) that indicates the potential of these units and could be used in many robotics applications.
The experimental results presented in this paper are all based on the custom smart beacon PCB (Figure~\ref{fig:circuit and camera}(a)).

The event camera used was the \textit{Prophesee Gen4} camera (HD, $1280 \times 720$ pixels).
All data were post-processed for the purposes of this paper.

\subsection{Transmission and Motion Rate Experiment}

\begin{figure}[t!]
\includegraphics[trim={20cm 0cm 15cm 10cm}, clip, width=0.62\linewidth]{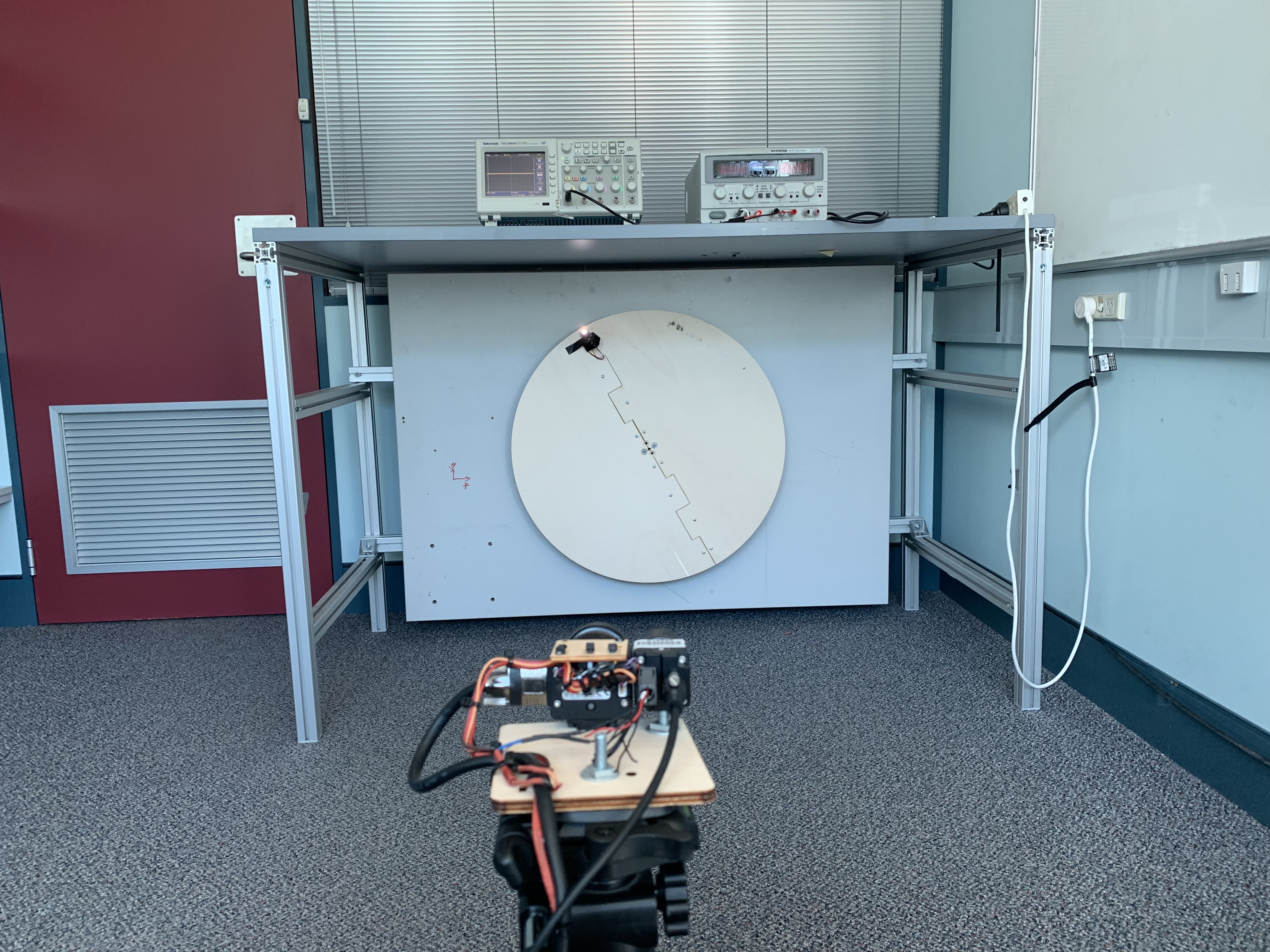}
\centering
\caption{
Experimental setup for spinning disk experiments.
The beacon is attached at the perimeter of a 700~mm diameter disk at a distance of 2~m from the camera.
The speed of the disk is controlled through the input frequency to a stepper motor.
The event camera is mounted alongside a classical machine vision camera (with a signal conditioning circuit to capture frame trigger signals of the machine vision cameras in the event stream) although only the event camera was used in the experiments documented in this paper.
}
\label{fig:Experimental_rig}
\end{figure}

In this experiment, the LED light source is mounted on a spinning disk as shown in Figure~\ref{fig:Experimental_rig}.
The beacon is programmed to transmit a short known ASCII code message at different transmission frequencies and for different speeds of the rotation of the disk.
Following~\cite{wang2022smart}, the message chosen was Sonnet 18 by William Shakespeare~\cite{shakespeare1989william}, ``Shall I compare thee to a summer's day?''
The message comprises 618 characters, including spaces and punctuation.
We transmit the Sonnet in a loop, with a 20-bit `ON-OFF' signal between two full messages that is not used in the bit or message error rate computations.

We quantitatively evaluate the communication system using two metrics: the message error rate (MER) and the bit error rate (BER).
The BER is the number of bit errors divided by the total number of transmitted bits before error correction, while the message error rate is evaluated on a per-packet basis after error correction: a packet is counted as correct only if all three ASCII characters are decoded correctly; otherwise, it is counted as erroneous.
We define the accuracy as the complement of the error rate.
So the transmission message accuracy rate is (1 - MER) and the bit accuracy rate is (1 - BER) in Tables~\ref{tab:packet_accuracy_static}--\ref{tab:packet_accuracy_spin}.
The accuracy is computed based on the first full sonnet received, a total of 618 characters (206 packets), or $32 \times 206 = 6592$~bits.

The results presented are dependent on the environmental lighting conditions and LED light brightness used.
We control for these factors by undertaking all experiments in similar indoor daylight lighting conditions ensuring that the internal setting for background light and sensitivity for the Prophesee Gen4 camera are similar.
The pulse duration is set to 50\% of the period of the transmission rate.
For this choice, the rising edge MFM modulation scheme is illuminated for 37.5\% of the time regardless of the transmission rate, and the average beacon brightness is constant through all the experiments.
We also ensure that all experiments are undertaken with the same LED and a fully charged battery.
The first experiment was undertaken with a static smart beacon.
That is, the disk is not spinning and the LED is stationary in the image.
Table~\ref{tab:packet_accuracy_static} presents results for a range of transmission rates and associated LED pulse durations.
In the second experiment, the disk was rotated at three different speeds: 3.5, 6.5, and 9.2~kpx/s.
Note that even the slowest setting is still a very significant optical flow (3.5~kpx/s is roughly 175~px/frame at 20~Hz frame rate) for traditional computer vision systems.
In Table~\ref{tab:packet_accuracy_spin}, the bit and message accuracies are evaluated for five transmission rates at the three different speeds.

For a static beacon, our algorithm achieves 100\% bit accuracy (and consequently message accuracy) up to 12.5~kb/s.
Beyond this rate, the communication accuracy degrades quickly as we approach the hardware limitations of the event camera and the intensity signal cannot be accurately reconstructed (cf.~Figure~\ref{fig:static_10kbps_16kbps}).
The bit accuracy degradation is relatively smooth as the signal-to-noise ratio decreases with increasing bandwidth (cf.~Table~\ref{tab:packet_accuracy_static}).
One can note that increasing optical flow does degrade the bit accuracy by comparing the 12.5~kb/s and higher transmission rate at different speeds in Table~\ref{tab:packet_accuracy_spin}.
Degradation in bit accuracy with increased optical flow appears roughly linear once errors begin to occur: $\sim$10--20\% decrease in bit accuracy for each increase of $\sim$3~kpx/s in optical flow for transmission rates above $\sim$12.5~kb/s.
However, it is only once the limits of transmission are reached that the motion of the target degrades bit accuracy, and we measure accuracy rates of 100\% for transmission rates up to 11.11~kb/s for all optical flow rates tested.

The message error rate is computed following packet decoding.
Because the Hamming (32,26) packet provides 1-bit error correction, the message error rate is robust to low bit error rates.
When the bit error rate increases to the point that two or more bit errors occur in a single packet the message accuracy rate shows aggressive degradation.
Characters with bit sequences containing repeated 1s such as ${\cdots}111{\cdots}$ are more likely to generate errors due to the leading edge detection failing (cf.~Figure~\ref{fig:static_10kbps_16kbps}).
Packets with more than one such character are more likely to cause packet loss than those with a single repeated bit sequence due to the error correction.
It is likely that careful design of dedicated line and channel encoding schemes targeted to event camera optical communication could improve these results, however, it is unlikely that such improvement would be practically significant given that the failure of the proposed methodology occurs close to the limits of the hardware anyway (refractory period, timing jitter, \etc).

Using a brighter LED would improve the experimental results since the leading edge detection is dependent on the level of irradiance of the pixel in the camera.
An event camera with improved sensitivity and reduced refractory period would also improve the observed results, although such a camera would then generate more spurious events and the detection and tracking algorithm would need to deal with these events without becoming overloaded.
The \textit{Prophesee Gen4} camera is one of the industry-leading event cameras available at the moment and the LED beacons used were designed for fast response and a powerful LED.
We believe the results provided are indicative of the potential of the technology.

\begin{table}[h]
	\newcolumntype{P}[1]{>{\centering\arraybackslash}p{#1}}
	\centering
	\caption{Bit and message accuracy for a static beacon. The LED pulse duration is dwell time for illumination of the beacon programmed in the STM32 for each LED voltage state (L or H) in the line encoded signal.}
	\begin{tabular}{|P{0.18\columnwidth}|P{0.18\columnwidth}|P{0.15\columnwidth}|P{0.18\columnwidth}|} \hline
	\textbf{LED Pulse Duration ($\mu s$)} & \textbf{Transmission Rate (kb/s)} & \textbf{Bit Accuracy, 1-BER,  (\%)} & \textbf{Message Accuracy,\makebox[1mm]{} 1-MER,\makebox[1mm]{}  (\%)} \\
	\hline
	60 & 8.33  & 100.00 & 100.00 \\ \hline
	55 & 9.09  & 100.00 & 100.00 \\ \hline
	50 & 10.00 & 100.00 & 100.00 \\ \hline
	45 & 11.11 & 100.00 & 100.00 \\ \hline
	40 & 12.50 & 100.00 & 100.00 \\ \hline
	35 & 14.29 & 94.86  & 92.23 \\  \hline
	30 & 16.67 & 62.64  & 43.69 \\  \hline
	25 & 20.00 & 28.25  & 0.50 \\ \hline
	20 & 25.00 & 0.00 & 0.00 \\
	\hline
	\end{tabular}
	\label{tab:packet_accuracy_static}
\end{table}

\begin{table}[ht]
	\newcolumntype{P}[1]{>{\centering\arraybackslash}p{#1}}

	\caption{Bit and message accuracy for a moving beacon. The speed (pixels per second) is that of the beacon on the image plane. For each transmission rate, three speeds were evaluated. }
	\centering
	\begin{tabular}{|P{0.16\columnwidth}|P{0.18\columnwidth}|P{0.08\columnwidth}|P{0.15\columnwidth}|P{0.15\columnwidth}|}
			\hline
	\textbf{LED Pulse Duration ($\mu s$)} & \textbf{Transmission Rate (kb/s)} & \textbf{Speed (px/s)} & \textbf{Bit Accuracy, 1-BER, (\%)} & \textbf{Message Accuracy, 1-MER, (\%)} \\
	\hline
	\multirow{3}{*}{50} & \multirow{3}{*}{10} & 3.5k &  100.00 & 100.00 \\
						&					  & 6.5k &  100.00 & 100.00 \\
						&					  & 9.2k &  100.00 & 100.00 \\
	\hline
	\multirow{3}{*}{45} & \multirow{3}{*}{11.11} & 3.5k & 100.00 & 100.00 \\
						&						 & 6.5k & 100.00 & 100.00 \\
						&						 & 9.2k & 100.00 & 100.00 \\
	\hline
	\multirow{3}{*}{40} & \multirow{3}{*}{12.5} & 3.5k & 100.00 & 100.00 \\
						&						& 6.5k & 97.69  & 96.12 \\
						&						& 9.2k & 78.63  & 67.48\\
	\hline
	\multirow{3}{*}{35} & \multirow{3}{*}{14.29} & 3.5k & 95.07 & 91.75 \\
						&					     & 6.5k & 81.08 & 71.84 \\
						&					     & 9.2k & 61.64 & 46.12 \\
	\hline
	\multirow{3}{*}{30} & \multirow{3}{*}{16.67} & 3.5k & 65.14 & 40.78 \\
						&					     & 6.5k & 54.19 & 27.18 \\
						&					     & 9.2k & 47.33 & 18.93 \\
	\hline
	\end{tabular}
	\label{tab:packet_accuracy_spin}
\end{table}

\subsection{Detection Experiment}
\label{sec:Detection_Experiment}

\begin{figure}
    \centering
    \subfloat[]{\label{subfig:indoor_smart_city_a} \includegraphics[width=0.48\columnwidth]{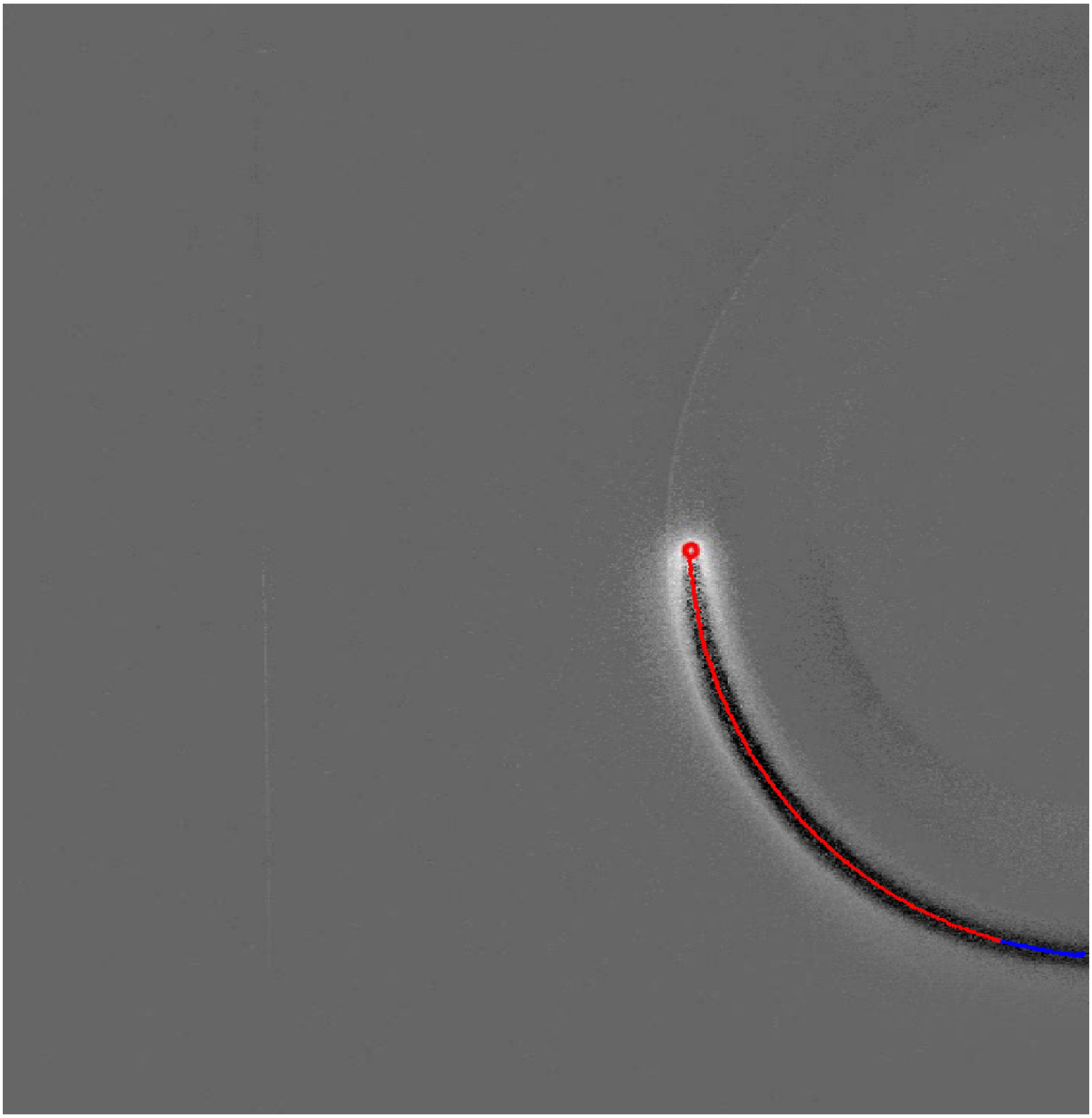}}
    \subfloat[]{\label{subfig:indoor_smart_city_b} \includegraphics[width=0.48\columnwidth]{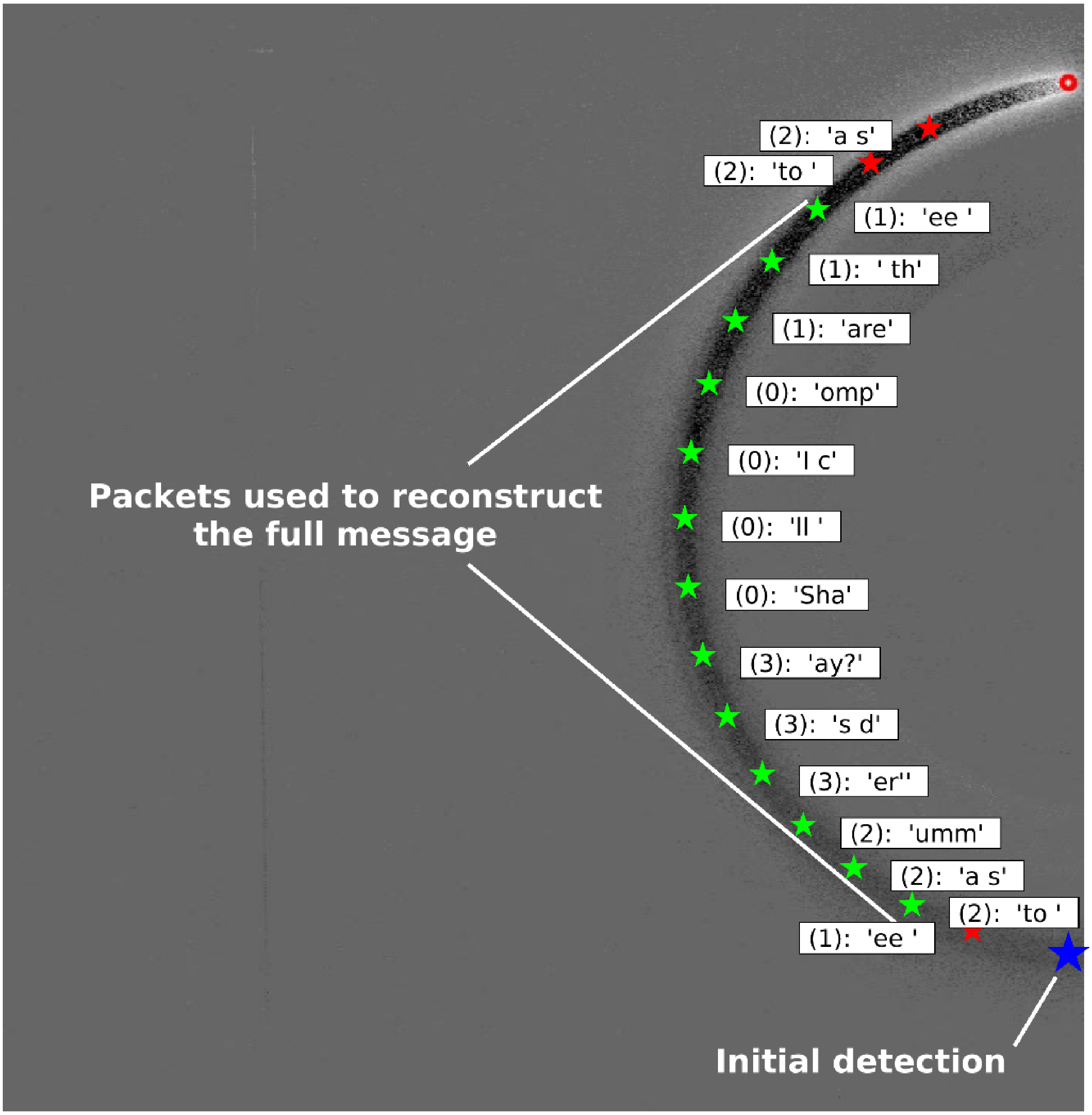}}
	\caption{
	Detection and tracking experiment.
	The beacon is moving at 12.5~kpx/s and enters and leaves the camera Field of View (FoV).
	(a) On entering the FoV, the beacon is identified as a candidate within 4 pixels (blue star), validated within 5~ms (red star), and tracked reliably through its trajectory.
	(b) Packets used to reconstruct the full message are shown (green stars) with the corresponding group ID and ASCII characters.
	The group ID allows the message to be reassembled even when the first packet decoded is partway through the message.
	Red stars at the top of the image indicate repeated data.
	To read the encoded message, start reading from group (0): `Sha' and read upwards in the direction of motion, then return to the first packet decoded once repeated data is received.
	}
	\label{fig:smart_city_indoor}
\end{figure}

In this experiment we demonstrate the capability of the system to quickly detect and track a moving beacon, and then decode a short message.
The experiment uses a smart beacon attached to the spinning disk as described in the previous section (cf.~Figure~\ref{fig:Experimental_rig}).
Unlike the previous experiment, the event camera captures only half of the trajectory, causing the beacon to rapidly enter and exit the field of view (cf.~Figure~\ref{fig:smart_city_indoor}).
No prediction was made for the beacon entry point, the algorithm relied purely on detection of a flickering light source, tracking of the light source, and then decoding.
The event stream was first captured and then post processed.

In this experiment, we transmitted a short message of 39 ASCII characters using 13 packets.
The packets are grouped into four blocks indexed by the 2-bit message type field, which we refer to as the \emph{group ID} when it is used in this role.
The first block of four packets contains the ASCII characters \{Sha\} - \{ll \} - \{I c\} - \{omp\} comprising the first few letters of the first line of Shakespeare's sonnet.
The remaining blocks contain the subsequent characters.
Separating the message into blocks indexed by the group ID allows the decoder to start capturing at any point in the transmission rather than waiting for the beginning of the message (green stars, Figure~\ref{fig:smart_city_indoor}).
After the full message has been received, the blocks can be reordered and the complete message reassembled.

The wheel was set to rotate at 6.8~rev/s, corresponding to a speed of 12.5~kpx/s in the image (Figure~\ref{fig:smart_city_indoor}).
Our detection algorithm detects the track within 4 pixels of it entering the field of view and validates the target within 5~ms, as seen in the bottom right of Figure~\ref{subfig:indoor_smart_city_b}.
The detection (blue star in Figure~\ref{subfig:indoor_smart_city_b}) is almost immediate, while the validation (red star in Figure~\ref{subfig:indoor_smart_city_b}) requires a full packet to be received.
With a transmission rate of 10~kb/s, each packet is transmitted in 3.55~ms, and in this example the first start delimiter occurred 2.8~ms after the beacon was detected.
In the example shown, we decoded 16 packets (48 characters) within 53.2~ms, achieving 100\% message accuracy.
Using the group ID, a full message of 13 packets (green stars in Figure~\ref{subfig:indoor_smart_city_b}) is decoded in 42.6~ms.
The additional packets (shown in red at the top right of Figure~\ref{subfig:indoor_smart_city_b}) can be discarded once the full message has been received.
The experiment was repeated multiple times with identical results.

\section{Smart City Localised Communications}
\label{sec:smart_city}

Many smart city digital interfaces need to be unobtrusive to the end user and spatially localised.
Examples include exchanging local information on speed, braking, \etc, between moving vehicles, providing local navigation information to pedestrians in buildings and complex built environments, providing tourist information to visitors in a city, enabling ticket entry to premises, making electronic payments, monitoring usage of infrastructure, \etc
Optical communication channels based on flickering LEDs and event cameras have the potential to provide a powerful technology for smart city digital interfaces.
Optical communications channels (through beacon tracking) have the advantage of providing information on where the agent that provided the information was located, which car provided the speed information, which person provided validation of ticket entry, \etc
Safety and security of data links is also a critical consideration in future applications; optical communication channels cannot be ``tapped'' easily since they require line of sight and this adds a level of security against eavesdropping and fraud.
LED lights are used widely in daily life, such as indoor and outdoor lighting, car lights and mobile phones, and are ideal for optical communication channels.
Infrared LEDs are not visible to human eyes and could be used for such applications without creating unwanted (human visible) light pollution as well.
The tools needed to encode information in an LED light source can easily be integrated into the hardware design, and are often already present in many cases.
Event camera technology is fundamentally the same as standard CMOS camera technology and event camera chips and processing units will become progressively cheaper and be integrated into consumer electronics over the next few years enabling such applications.

In this demonstration, a smart beacon is mounted on a cyclist's bicycle to provide information about the cyclist as they ride past a sensor.
A fixed camera is placed by the side of the path, and detects, tracks and decodes the information from the beacon as the bicycle passes the camera.
This scenario could be used, for example, in bicycle races where the cyclist passes through a gate to finish the race and is timed and identified as they pass.
More generally, any vehicle could be equipped with such a beacon, to provide ID information on entering parking areas, driving down reserved roads, or any other monitoring scenario.
The optical speed of the beacon is available from the track data, information that may also be useful depending on the application.

\begin{figure}[t!]
	\newcolumntype{P}[1]{>{\centering\arraybackslash}p{#1}}
	\centering
		\includegraphics[width=1\linewidth]{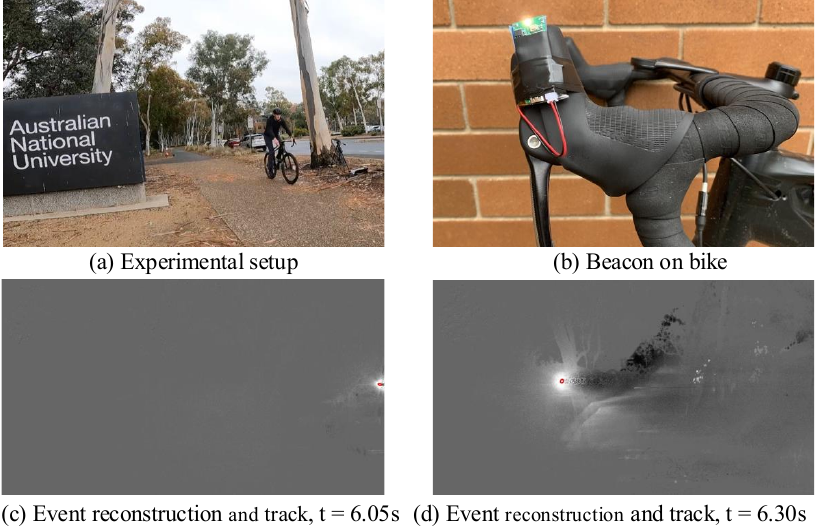}
	\caption{\label{fig:exp_bike}
		\textbf{Example of the smart city application.}
		(a) shows the experimental environment, where a bike rider passes by an event camera positioned at the exit of the university campus.
		(b) illustrates a smart beacon mounted on the bike's handlebar, transmitting the rider's information as they pass the camera.
		(c)--(d) show the event-based reconstruction and the tracked LED beacon marked in red.
		The beacon first enters the frame at $t=6.011$~s and is validated by $t=6.019$~s (within 8~ms).}
\end{figure}

The demonstration is shown in Figure~\ref{fig:exp_bike}(a-b), with the reconstructed event image and the tracked beacon marked in red shown in Figure~\ref{fig:exp_bike}(c-d).
The beacon encodes information about the cyclist, for example, in a bicycle race, the typical information would be the cyclist's name, competitor ID and category, using 16 packets (48 ASCII characters).
We use the same grouping algorithm discussed in \S\ref{sec:Detection_Experiment} to divide the message into four indexed groups to overcome the requirement to wait for the start of the message before beginning to decode the message.

During the experiment, the beacon transmitted at 5~kb/s and was observed for 327~ms as the cyclist passed.
In this time, 46 packets (138 ASCII characters) were decoded, providing approximately three full messages.
Of the 46 packets received, three packet errors occurred.
The first was single-bit error in a packet that was successfully corrected using the Hamming code.
In the second case, the start delimiter was not correctly identified leading to the loss of a whole packet.
The start delimiter for the following packet was correctly identified, but this packet had at least two bit errors and was discarded.
This indicates a short period of high noise in the data signal.
The missing data was recovered from redundant packets available in the repeated message transmission.
The significantly higher error rates observed in the real world scenario compared to the laboratory experimental scenario are indicative of the more complex visual situation, full daylight and more background events due to the image of the bicycle moving through the scene.
This impacted on the overall signal-to-noise ratio of the received signal degrading performance.
Even though the error rates were higher, we demonstrated reliable communication at 5~kb/s under fast motion and a complex background and reconstructed the message with 100\% accuracy.
The real world scenario demonstrates both the potential of the technology as well as the importance of experimental tuning of the system in real world scenarios.

\section{3D Optical Tracking and Reconstruction}
\label{sec:3D_optical_tracking_and_recon}
Optical Motion Capture (MoCap) systems observe a moving beacon (or feature) from multiple cameras and triangulate to recover its 3D position and motion \cite{Hartley03book}.
Because even small timing offsets can produce substantial spatial errors, accurate temporal alignment between cameras is essential, particularly for fast-moving targets.
For example, optical flow of 10~kpx/s captured with a 1~ms time latency/jitter corresponds to errors of up to 10 pixels, significantly impacting the spatial accuracy of reconstruction.
To overcome this limitation, conventional motion-capture system cameras are typically connected through a physical cabled interface used both for synchronisation and data transfer.
Such infrastructure is costly, time consuming to deploy, and difficult to scale.

The ATOC pipeline can be exploited to provide a powerful 3D visual motion capture system capable of very high temporal and spatial tracking resolution without any requirement for physical cabling.
The key idea is to deploy smart LED beacons that repeatedly transmit short messages containing an ID and packet counter.
A suite of independent event cameras runs the ATOC pipeline on local embedded compute, detecting and tracking each of the visible smart beacons.
They then decode the beacon message and use the beacon ID and packet counter to establish global time synchronisation locally without the requirement to externally synchronise the cameras.
The local camera then transmits the beacon ID, optical position, optical velocity, and global synchronisation information to a central computer, possibly over Wi-Fi (although in the present experiments the data is stored locally and post-processed). There, the beacon tracks are collated and the 3D position of the beacon is computed by triangulation.
The system is simple to configure, relatively cheap (depending on the cost of event cameras), easy to scale (each new camera is independent), and leads to high-performance tracking.
Using only two cameras, we demonstrate cross-camera synchronisation to within 4.54~$\mu$s and 3D reconstruction with a precision of 2~mm at a rate of 4~kHz, comparable to high-end state-of-the-art visual MoCap systems.
Spatial precision could be improved by adding further cameras.

\subsection{Beacon Encoding}
\label{sec:beacon_encoding}

We use a single packet of the Hamming (32,26) code as the message for each beacon to transmit.
One byte of data is dedicated to an 8-bit identification code for each beacon (allowing up to 256 individual beacons to be present in the same environment).
This allows us to differentiate beacons that operate in close proximity and re-identify beacons quickly after occlusion.
Two bytes of data are dedicated to a 16-bit packet counter.
The 2-bit message type field is not used in this application but could be used to transmit different types of message, for example, inertial data or diagnostics (battery status, \etc).
The packet counter allows for 65,536 packets to be individually labelled.
Transmitting at 10~kb/s, a packet takes 3.55~ms to transmit and the packet counter wraps every 232.7~s, or roughly every four minutes.
It follows that the local camera clocks only need to be synchronised to within this 232.7~s period using some external signal in order that the packet counter will be uniquely identified within the local processor.
In particular, local clocks could easily be synchronised over Wi-Fi, since a typical Wi-Fi synchronisation has a latency of at most 50~ms.
Similarly, if a calibration signal is used to initialise the experiment, then each camera can start their local clock when it first sees the calibration signal.
As long as all cameras see their first calibration signal within 230s then they will all be correctly calibrated.
For the experiments reported in this paper we used a calibration signal.

\begin{figure}[t]
	\centering
	\includegraphics[width=1\linewidth]{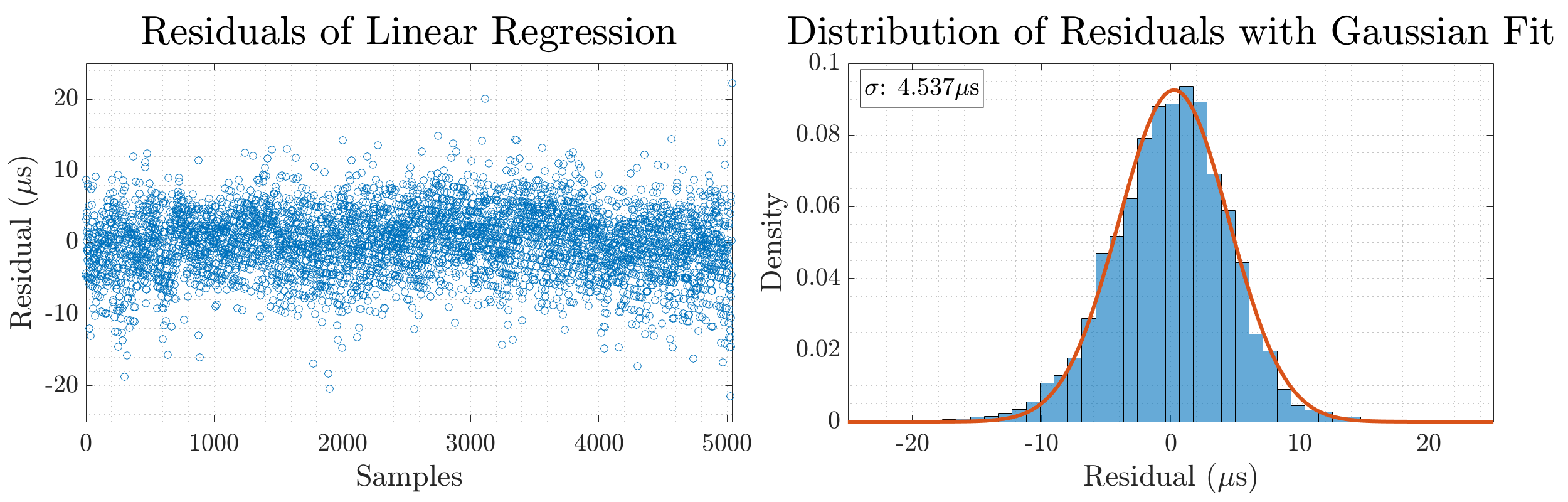}
	\caption{\label{fig:time_offset_distribution}
		\textbf{Cross-camera clock offset residuals.}
		Left: residuals of a linear fit to the measured time offset between the two event cameras, computed from beacon timestamps decoded simultaneously by both cameras, plotted against sample index.
		Right: histogram of the residuals with a fitted Gaussian distribution ($\sigma = 4.54~\mu$s), indicating the precision of the cross-camera temporal alignment.}
\end{figure}

The global packet counter is only used for correlation of data.
Accurate temporal synchronisation of the AEB track data is based on start delimiter leading edge timestamp detection.
The start delimiter leading edge is the most reliable edge detection for each packet (due to the LLLLH sequence) and it is the timestamp of this leading edge (in the local camera clock) that is used as the primary signal to associate the position and velocity of the AEB track state (computed in the local clock) to a specific global temporal event that can be correlated between cameras.
To quantify the precision of the temporal alignment at this timestamp, we undertook an experiment to estimate the error distribution associated with the start delimiter leading edge timestamp.
We model each local camera clock as a linear model with unknown time offset and clock drift (difference in clock rate).
In order to remove dependence on the beacon we do linear regression \emph{between the two cameras}, that is, we regress one camera's start delimiter leading edge timestamps against those of the other camera.
We fit a Gaussian distribution to the residuals of the linear fit of this regression as shown in Figure~\ref{fig:time_offset_distribution}.
The Gaussian has a standard deviation of $\sigma = 4.54~\mu$s.
Assuming the two cameras are stochastically independent, the variance of the residual is the sum of the individual variances in onset detection timestamp in each camera, giving a per-camera standard deviation of $\sigma = 3.21~\mu$s.
For a two-camera system, the two measurement jitter error $\sigma = 4.54~\mu$s can be directly used as an estimate of relative timestamp error.
Note that at 10~kpx/s optical flow, the timing error corresponds to a 0.05~px error, making this insignificant in the error analysis of the final 3D reconstruction.

\subsection{Intrinsic and Extrinsic Calibration}
\label{sec:calibration}

\begin{figure}[!h]
    \centering
    \subfloat[]{\label{subfig:checkerboard_a} \includegraphics[height=0.185\columnwidth]{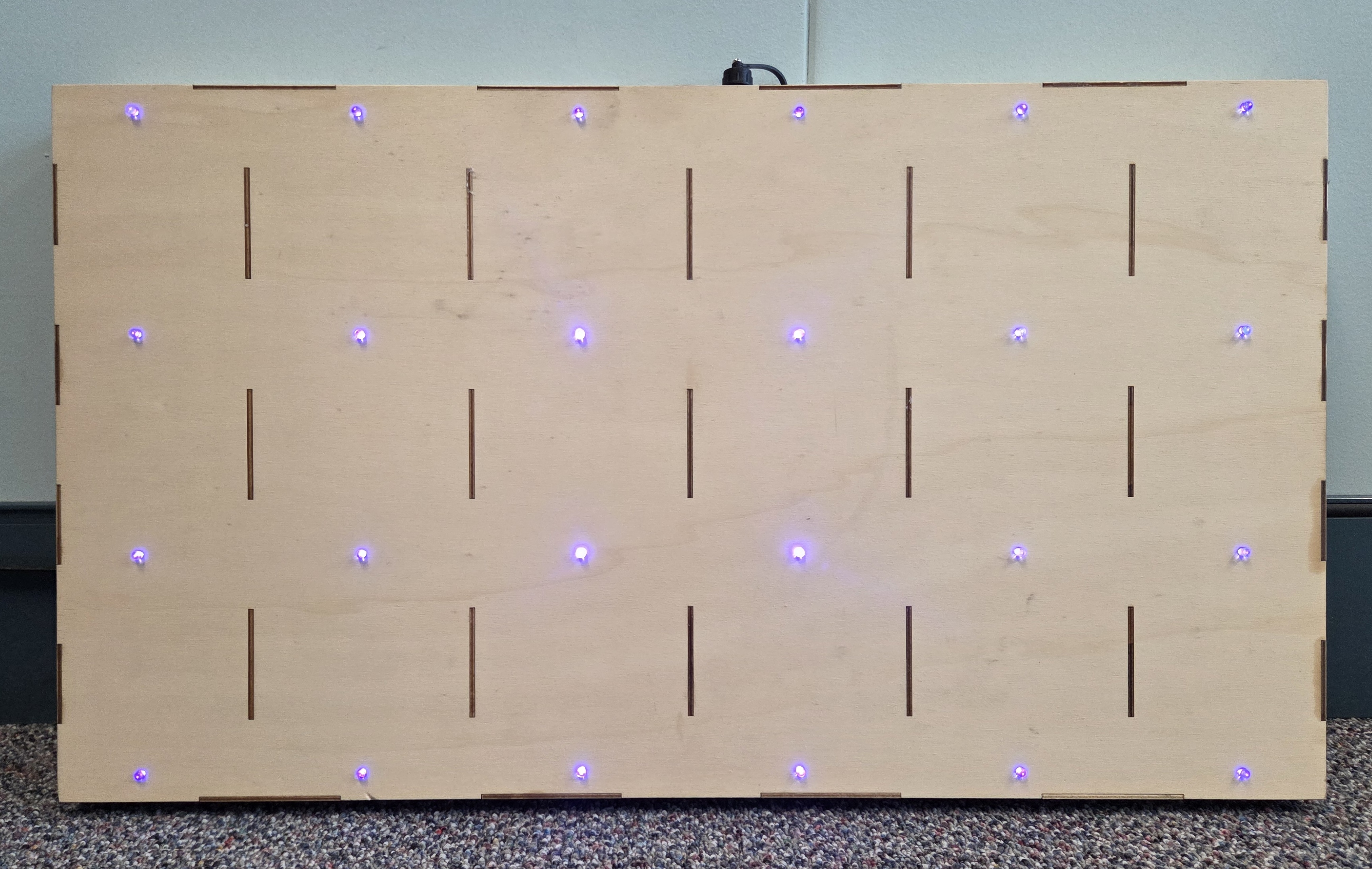}}
    \subfloat[]{\label{subfig:checkerboard_b} \includegraphics[height=0.185\columnwidth]{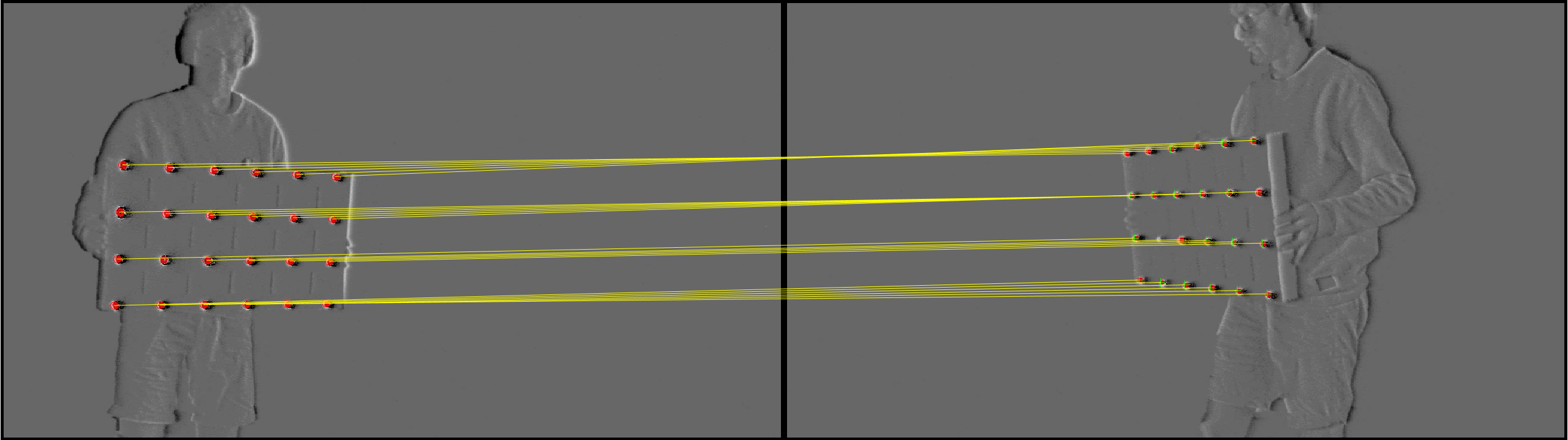}}
	\caption{\textbf{LED Calibration Board.} (a) Our calibration board with 24 LEDs, each transmitting a unique ID. (b) Feature point correspondence is simplified using the unique beacon IDs and the packet counter for temporal synchronisation.
	The image reconstruction serves only as a visualisation, camera calibration is performed directly using the AEB tracker coordinates.}
	\label{fig:checkerboard}
\end{figure}

Intrinsic and extrinsic calibration of the cameras is extremely important in 3D reconstruction for MoCap.
In order to calibrate the event cameras, we built a dedicated ``checkerboard'' calibration rig (cf.~Figure~\ref{fig:checkerboard}).
The rig has flickering LEDs located precisely in a laser cut grid of known dimensions.
We use the same data packet (with ID and packet counter) discussed in \S\ref{sec:beacon_encoding} and use the same timing and global synchronisation algorithm to synchronise AEB tracks of individual LEDs, both within a single camera (for intrinsic calibration) and between cameras (for extrinsic calibration).
There is no need to hold the checkerboard stationary while collecting calibration data since the time synchronisation is so precise.
A movie of the calibration sequence is provided in the supplementary video.
We select a sequence of times to sample the data stream to generate `frame-like' data for the standard checkerboard calibration routines.
In the present case, we used 100~ms (10~Hz) sampling of the data stream.
We then used MATLAB's \textit{Computer Vision Toolbox} to perform both the intrinsic and extrinsic calibration.

\subsection{Visual Motion Capture Experiment}
\label{sec:led_pyramid_reconstruction}
\begin{figure}[h!]
	\centering
	\includegraphics[width=0.98\linewidth]{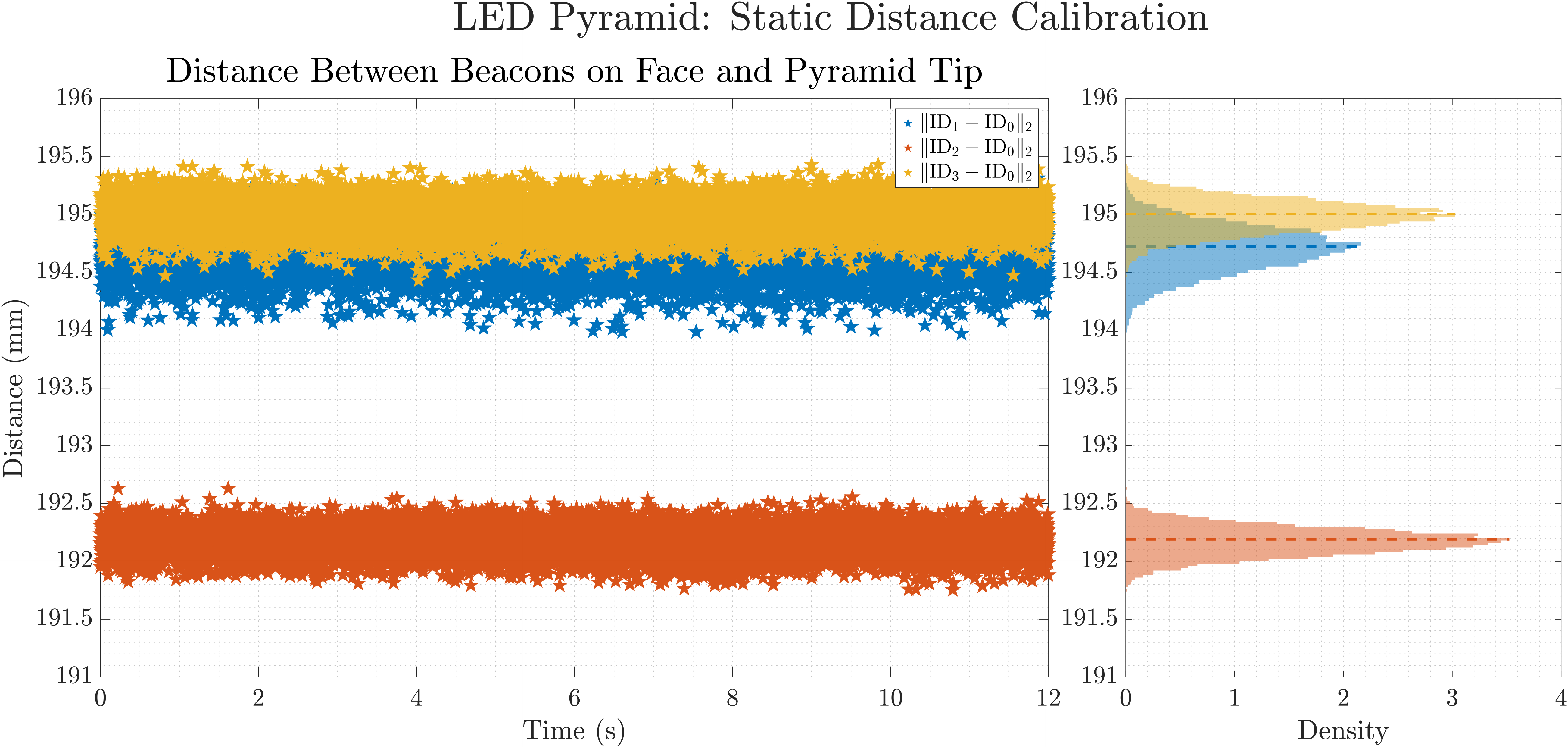}
	\caption{
	\textbf{Calibration of the pyramid baseline distances.} Because the beacons were hand-mounted, their true separations are measured before the dynamic experiments, with the pyramid held stationary in the centre of both camera fields of view. This calibrates the geometry of the rig; the intrinsic and extrinsic \emph{camera} calibration is a separate procedure, described in \S\ref{sec:calibration}.
	Left: the three distances from the apex beacon ID$_{0}$ to the face beacons ID$_{1}$, ID$_{2}$ and ID$_{3}$ over time. Right: the corresponding histograms, with the mean indicated by a dashed line.
	The sample means and standard deviations are listed in Table~\ref{tab:InterBeaconDistances} and serve as the reference distances for Figure~\ref{fig:3d_reconstruction_distances}.
	}\label{fig:led_pyramid}
\end{figure}

\begin{figure}[!h]
    \centering
    \subfloat[]{\label{subfig:led_pyramid_a} \includegraphics[width=0.4\columnwidth]{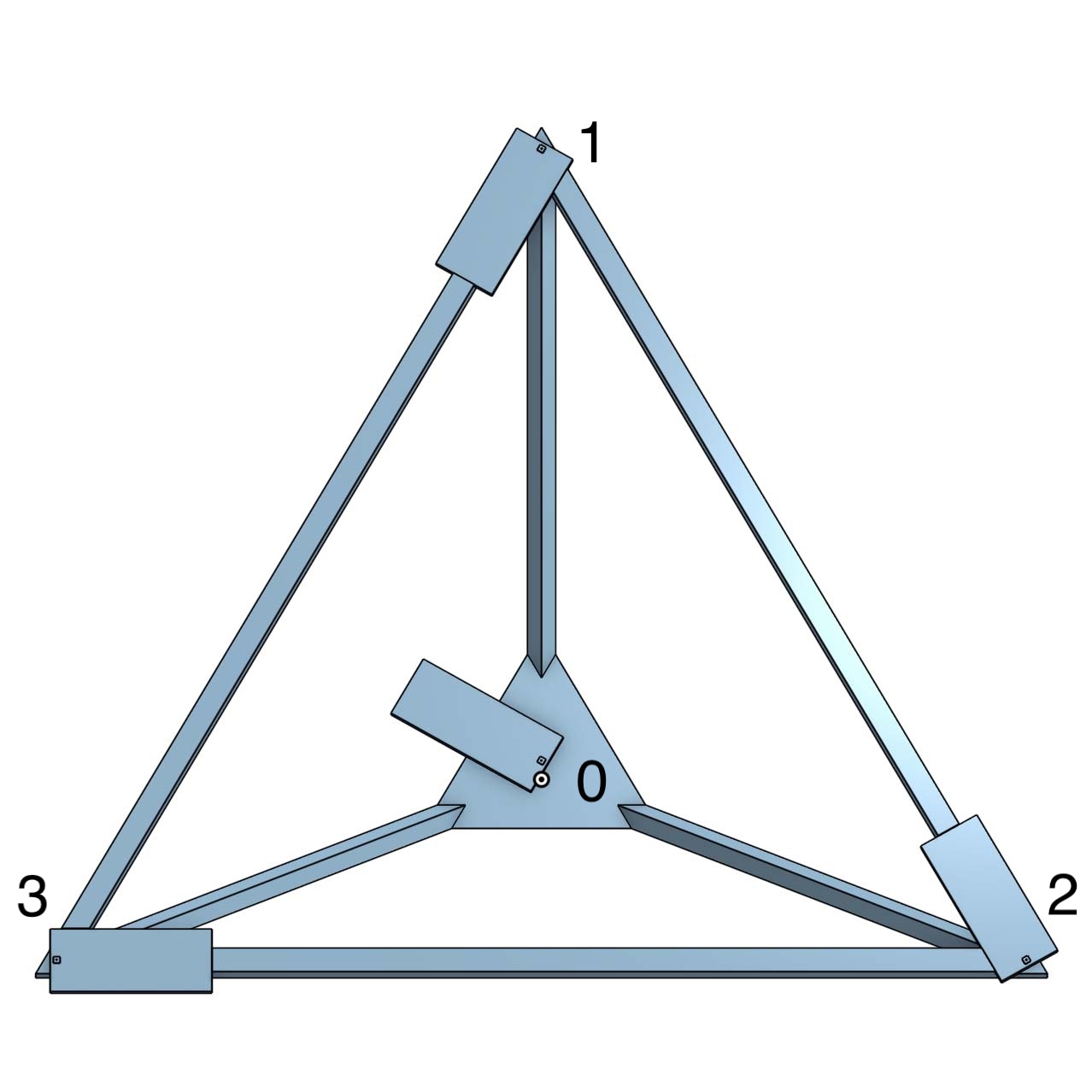}}
	\hspace{5mm}
    \subfloat[]{\label{subfig:led_pyramid_b} \includegraphics[width=0.4\columnwidth]{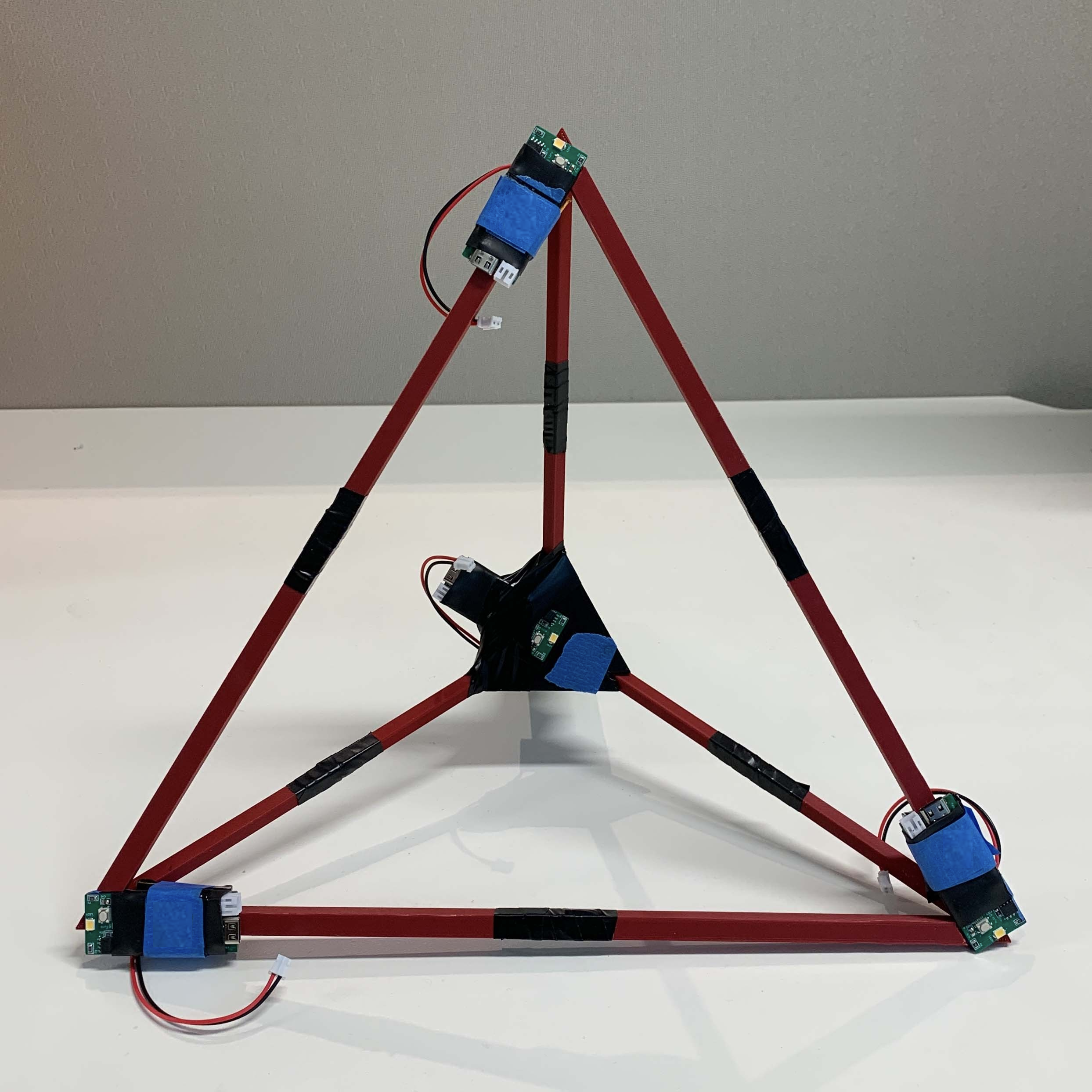}}
	\caption{Rigid 3D structure used to evaluate the precision of our 3D reconstruction. (a) The beacons and pyramid were modelled to the specified dimensions in CAD software, with the beacon IDs as shown. (b) The frame was 3D printed and the beacons were rigidly attached.
	The frame was printed in sections due to limitations in the 3D printer bed size, which introduced some structural offsets.}
	\label{fig:exp_led_pyramid}
\end{figure}

To characterise the performance of the 3D reconstruction, we constructed a rigid pyramid structure to mount four LED beacons in known positions (see Figure~\ref{fig:exp_led_pyramid}).
Three beacons (ID$_{1}$, ID$_{2}$, ID$_{3}$) are positioned on a planar face so that each pair of LEDs is roughly 300~mm apart.
The fourth beacon (ID$_{0}$) is positioned roughly 85~mm behind this plane.
This provides a full 3D structure to test relative depth reconstruction and removes the possibility that the homography structure of the planar calibration checkerboard biases the reconstruction results.
However, since we hand-mounted the beacons on the pyramid, it was necessary to first generate baseline data for the actual distances between beacons.
To estimate the 3D structure of the pyramid, we collected data with the pyramid stationary in the central field of view of each camera to minimise calibration errors (intrinsic and extrinsic).
The three distances from the apex beacon to the face beacons (ID$_{0}$-ID$_{1}$, ID$_{0}$-ID$_{2}$, ID$_{0}$-ID$_{3}$), and the corresponding histograms, are shown in Figure~\ref{fig:led_pyramid}.
The distribution for these distances is clearly Gaussian and we apply standard least squares inference to estimate the inter-beacon distances.

\begin{table}[h]
	\newcolumntype{P}[1]{>{\centering\arraybackslash}p{#1}}
	\centering
	\caption{Inter-beacon distances}
	\begin{tabular}{|P{0.18\columnwidth}|P{0.18\columnwidth}|P{0.18\columnwidth}|P{0.18\columnwidth}|} \hline
		\textbf{Beacons} & \textbf{Distance} & \textbf{Error} ($\pm 1\sigma$)  & \textbf{Color} \\ \hline
		ID$_{0}$ and ID$_{1}$ & 194.69~mm & 0.194~mm & Blue  \\ \hline
		ID$_{0}$ and ID$_{2}$ & 192.17~mm & 0.116~mm & Orange \\ \hline
		ID$_{0}$ and ID$_{3}$ & 194.99~mm & 0.135~mm & Yellow \\ \hline
	\end{tabular}
	\label{tab:InterBeaconDistances}
\end{table}

To evaluate the performance of the algorithm in dynamic motion we consider general rapid 3D motion of the pyramid, including both translation and rotation, obtained by waving the pyramid around in the cameras' field of view by hand.
Due to less optimal lighting and occasional occlusions, we decreased the transmission rate to 5~kb/s to ensure robust communication without packet loss.
This corresponds to a packet transmission time of 7.1~ms.

\begin{figure*}[!t]
	\centering
	\includegraphics[width=\textwidth]{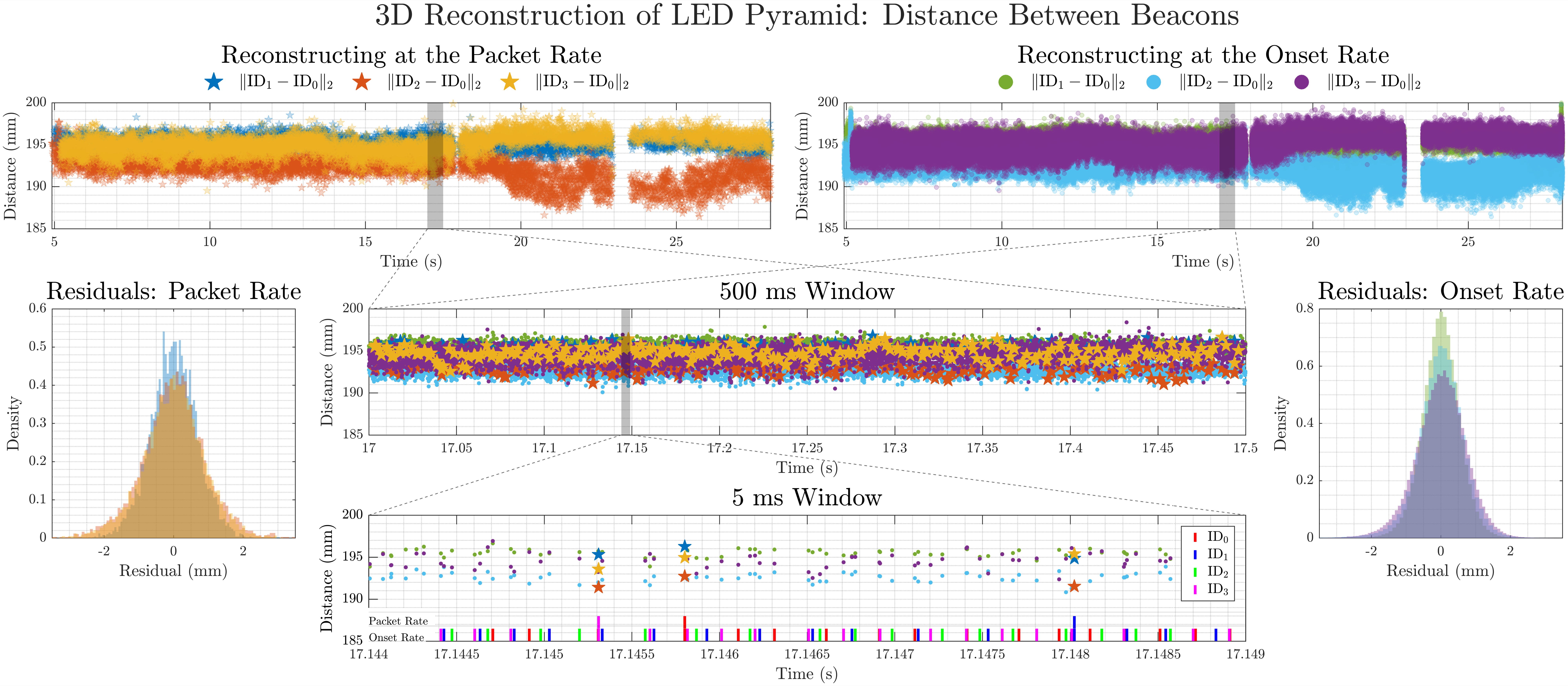}
	\caption{Evaluation of the spatial precision of 3D reconstruction using ATOC for beacon tracking and wireless time synchronisation. Four beacons are programmed to transmit an ID and a packet counter that are used to synchronise the two cameras.
	The distance between the apex beacon (ID$_{0}$) and each of the three face beacons (ID$_{1}$, ID$_{2}$, ID$_{3}$) of the pyramid structure (see Figure~\ref{fig:exp_led_pyramid}) is compared to the measured baseline distances (194.7, 192.2 and 195.0~mm; see Table~\ref{tab:InterBeaconDistances}).
	The distances are reconstructed when a full packet is received (packet rate, left) and at each matching onset point (onset rate, right).
	The two central plots show 500~ms and 5~ms time windows, respectively, highlighting the high temporal resolution our algorithm achieves. The residuals of the distance around the moving mean are shown in each histogram. }
	\label{fig:3d_reconstruction_distances}
\end{figure*}

Figure~\ref{fig:3d_reconstruction_distances} shows the Euclidean distances for the three beacon pairs considered.
The top row shows the estimates at the packet rate and onset rate, respectively.
The distance is shown by beacon pairs (e.g., ID$_0$ and ID$_1$) with the ID number also shown in Figure~\ref{fig:exp_led_pyramid}\subref{subfig:led_pyramid_a}.
The histograms show the residual distance estimates around the moving mean for the respective reconstruction rates.

There is slightly more variance in the 3D localisation observed (Figure~\ref{fig:3d_reconstruction_distances}) than was visible in the static baseline experiment (Figure~\ref{fig:led_pyramid}).
The largest disparity is in the period $\sim$20--25~s for ID$_{0}$ and ID$_{2}$ (orange and light blue, respectively).
This is during a period of very fast rotation of the pyramid relatively close to the edge of field of view of the cameras, and is most likely due to the constant acceleration assumption in the AEB (that tends to throw circular track motion outwards around the centre of rotation) combined with some calibration error.
Having noted this, the maximum deviation from the mean is $\sim$10~mm.
The overall precision of the scheme is of the order of 2~mm.
For a focal length of $1650$ pixels and a distance of $3$~m as used in the experiments, one pixel corresponds to $\sim$1.8~mm in the real world.
As such, we are reaching the physical limitations of the sensor and lens combination.
A higher focal length lens would increase the spatial resolution, but it would significantly reduce the field of view which is already limited in the current configuration.
The \textit{Prophesee Gen4} camera (HD, $1280 \times 720$ pixels) is the highest-resolution commercially available event camera, so there are no immediate alternatives that would improve performance.
We believe that the resulting system is close to the limits of precision that can be achieved by existing technology and is comparable to state-of-the-art MoCap systems.
Additional cameras would of course improve the precision of the 3D localisation of the point as well as add to the available workspace.

To highlight the high temporal resolution that can be achieved using ATOC, we perform 3D reconstruction at both the packet rate (reconstruct after each full packet is received) and the onset rate (reconstruct at each leading edge detection).
The two central plots progressively zoom in on the time axis, as indicated by the shaded bars and dashed lines.
The duration is included in the title of each of the subplots.
Here, the two reconstruction rates are overlaid, and the colours and markers follow from the top plots.
Reconstructing at the packet rate can be considered the base rate of our algorithm.
At a transmission rate of 5~kb/s, the packet rate is approximately 141~Hz for a single beacon.
This is comparable to a standard commercial motion capture system.
However, when multiple beacons are used, a decoded timestamp from any beacon can be used to synchronise the two event streams and perform triangulation.
In general, beacons are not synchronised and the effective base rate will vary between 141~Hz and 563~Hz (for four beacons) depending on the distribution of the onsets.
This highlights one of the unusual aspects of the proposed system, that the output can be asynchronous, that is, samples do not occur at specific times.
Indeed, ATOC naturally allows for even higher asynchronous reconstruction rates without interpolation by using all the detected onset points of a message, rather than just the start delimiter.
The onset points from a beacon are common (for the same packet), and hence this can be used to provide finer temporal alignment.
For a 32-bit packet encoded using MFM, there is a minimum of 16 and a maximum of 32 onsets per packet (average of 27 onsets per packet in the experiment), which provides more than an order of magnitude increase in the reconstruction rate for a single beacon ($\sim$4~kHz).
As with the packet rate, this rate increases for multiple beacons.
Note that at the reduced transmission rate of 5~kb/s all of the leading edge detections are relatively precise and the temporal synchronisation will be of a similar order of magnitude to that discussed in \S\ref{sec:beacon_encoding}.
In short, the ATOC system has sample rates well above those of most common visual MoCap systems.

In Figure~\ref{fig:3d_reconstruction_distances}, the 500~ms window begins to show the difference in the reconstruction rate.
In this interval, the packet rate generates 241 reconstructed points and the onset rate generates 6534 points.
This is further highlighted in the 5~ms window (bottom), where there are 3 reconstructed points at the packet rate and 76 points at the onset rate.
The colours indicate which beacon generated the update for the packet and onset rate, respectively.
As the packet and onset points provide an aligned marker in each event stream, all beacons can be sampled (propagated through the motion model if necessary) when an individual packet or onset is detected, so that all three distances are displayed at every reconstruction instant.
However, the 3D reconstruction remains asynchronous in time and is triggered by the decoded signals.
Reconstructing at the packet rate is comparable (temporally) to a traditional motion capture system; however, reconstructing at the onset rate provides a significant increase in performance above and beyond all but the very fastest existing motion capture systems.

\begin{figure*}[t!]
	\centering
	\begin{tabular}{ccc}
		\includegraphics[width=0.31\linewidth]{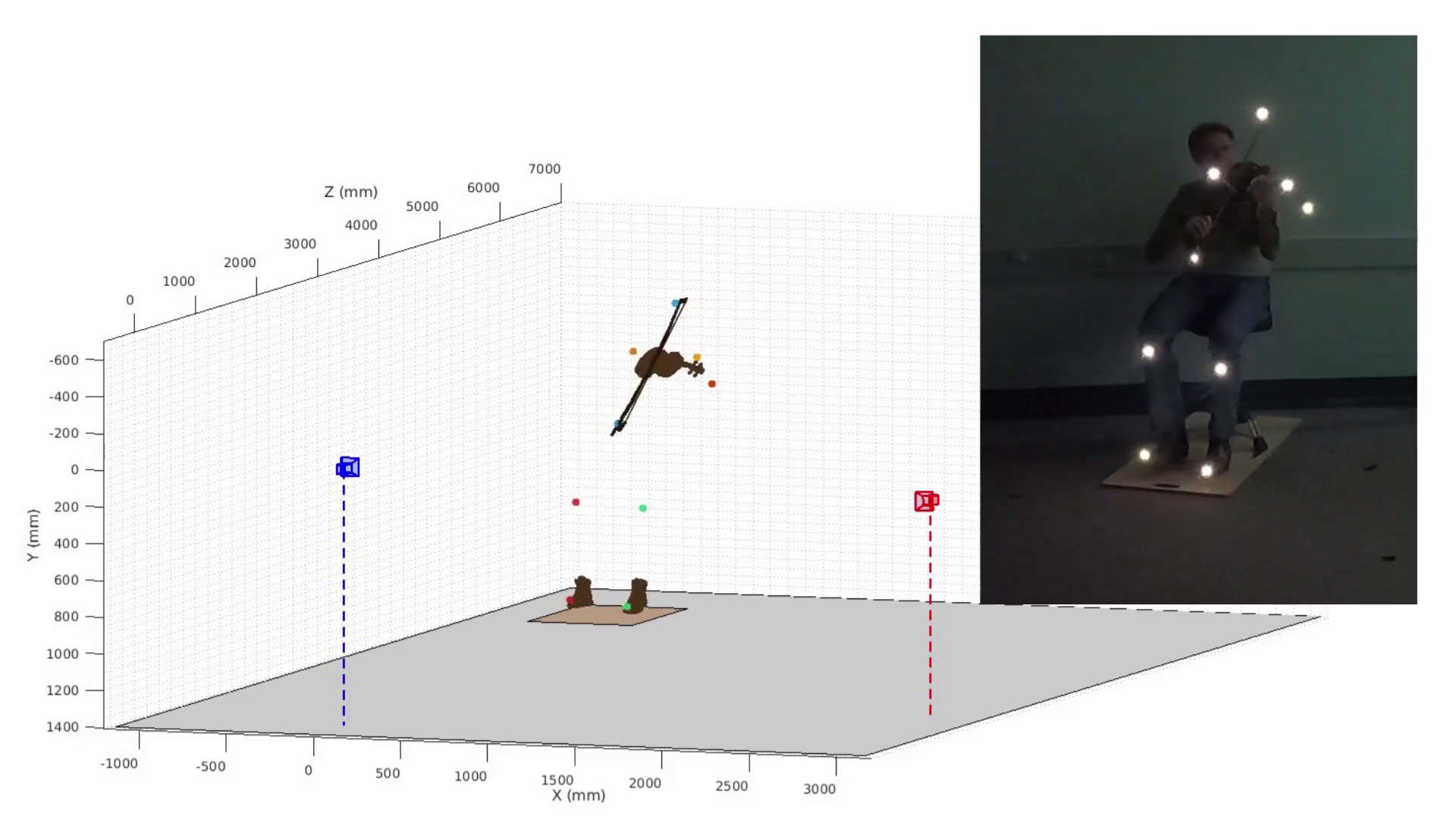} &
		\includegraphics[width=0.31\linewidth]{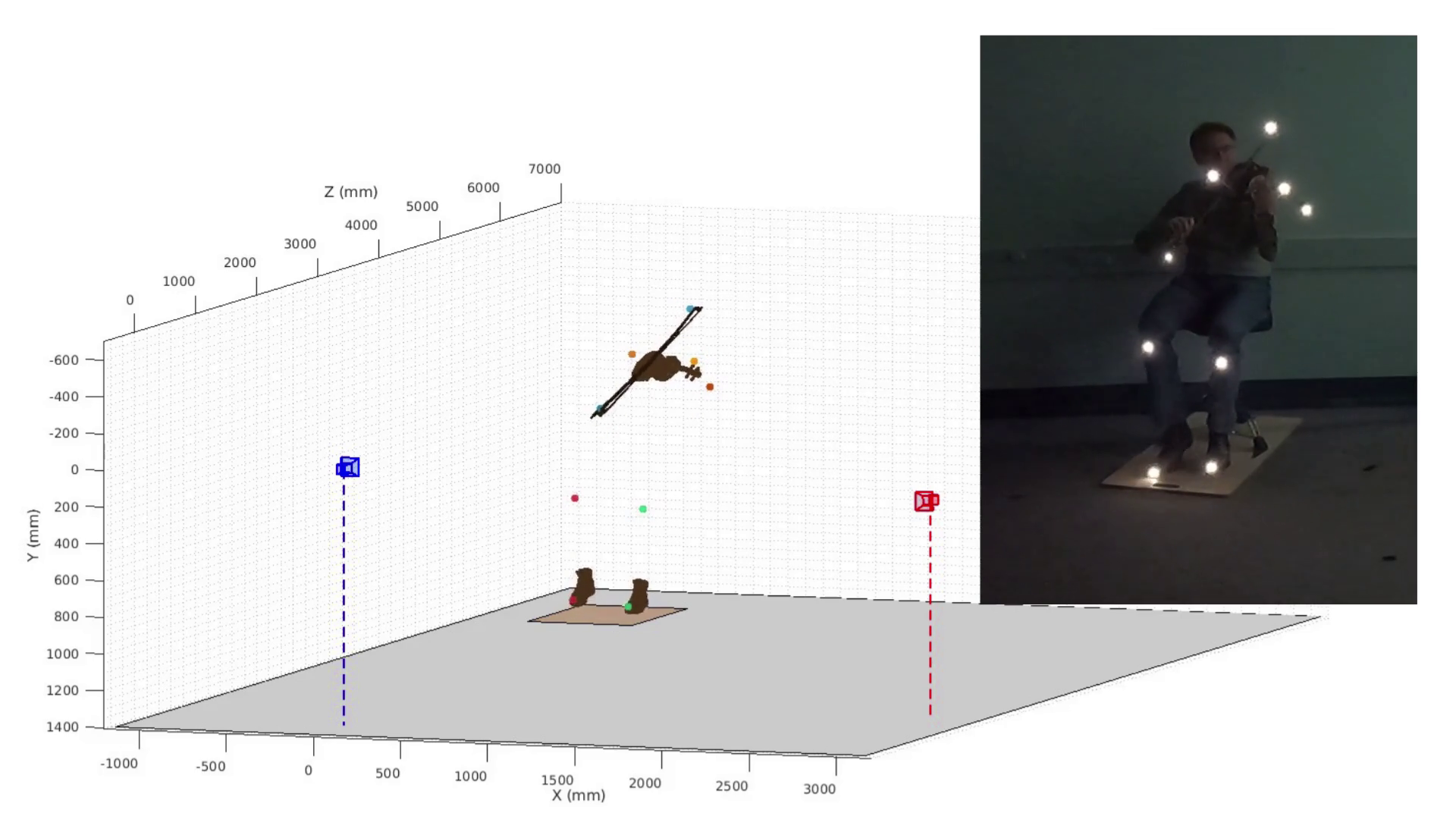} &
		\includegraphics[width=0.31\linewidth]{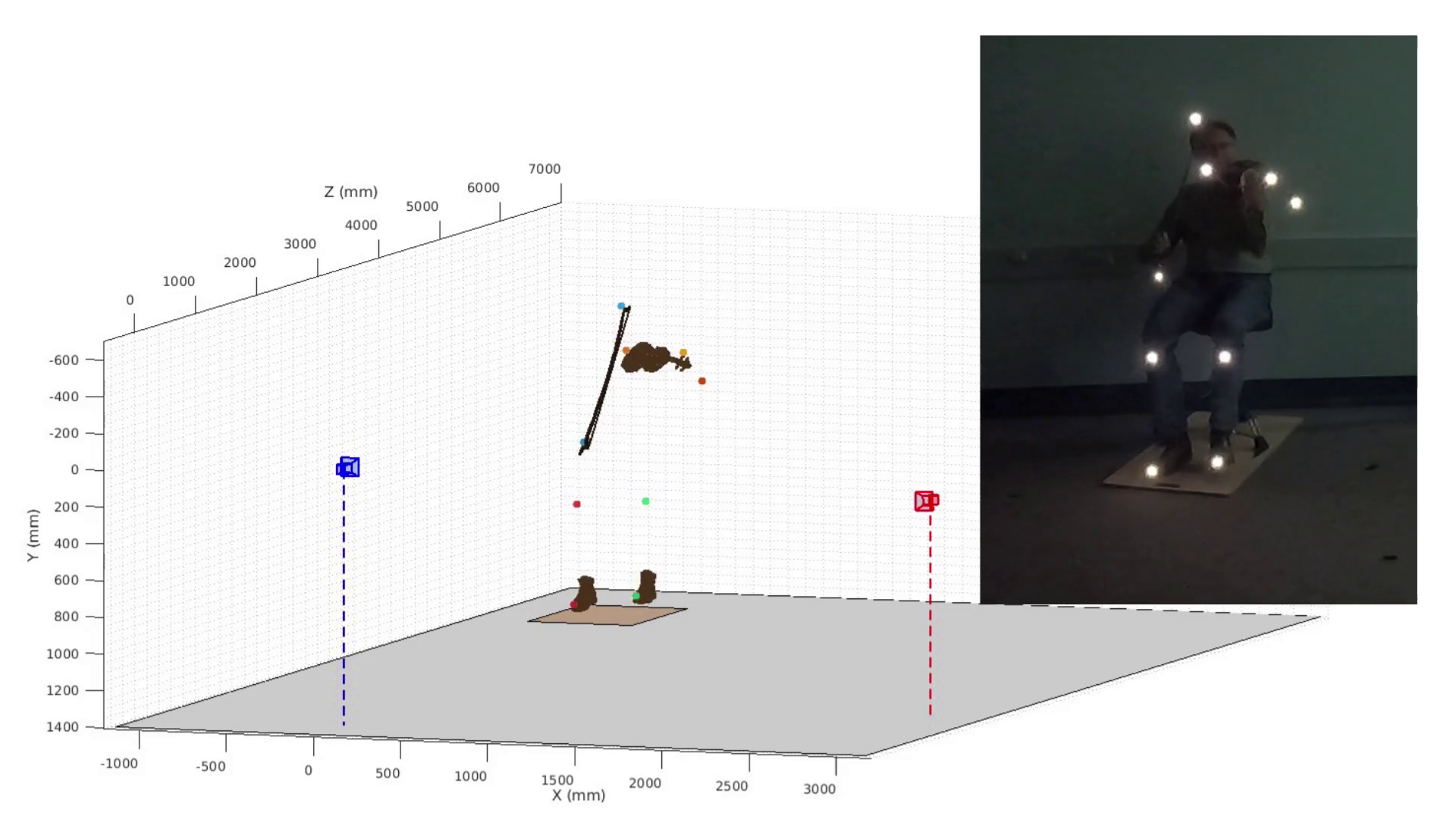} \\
		$t = 5.4$~s & $t = 8.2$~s & $t = 37.2$~s
	\end{tabular}
	\caption{
		\textbf{An example of 3D reconstruction results.}
		The two event cameras are shown as blue and red camera frusta in the 3D scene.
		A reference RGB frame is displayed to the right, showing a person playing the fiddle with nine LED beacons attached to the instrument and body.
		In the 3D reconstruction, the beacons are colour-coded by the part they are attached to.
		The 3D models for the violin, bow and boots were sourced from \cite{CAD_model_violin,CAD_model_bow, CAD_model_boot}, respectively.
	}\label{fig:3d-demo}
\end{figure*}

\begin{figure}
	\centering
	\begin{tabular}{c}
		\\	\includegraphics[width=1\linewidth]{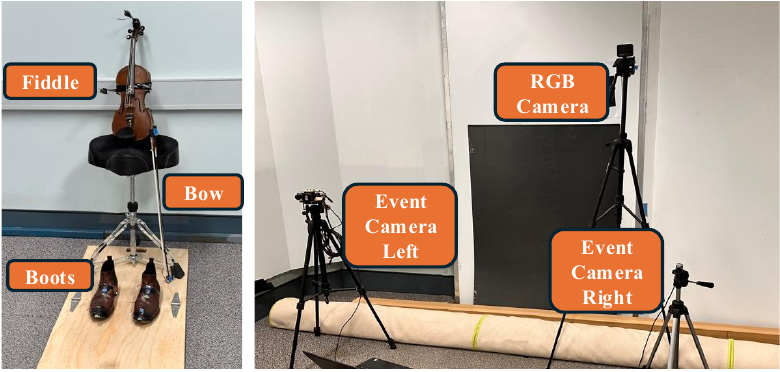}
	\end{tabular}
	\caption{
		\textbf{Experimental setup for 3D reconstruction using the ATOC algorithm.}
		Left: A fiddle, bow, and boots, attached with 9 programmed LED beacons.
		Right: Stereo event cameras are used to capture data for 3D reconstruction.
		An RGB camera records reference video, which is not used in the ATOC pipeline.
	}\label{fig:3d-setup}
\end{figure}

\begin{figure}[h!]
	\centering
	\begin{tabular}{c}
		\\	\includegraphics[width=1\linewidth]{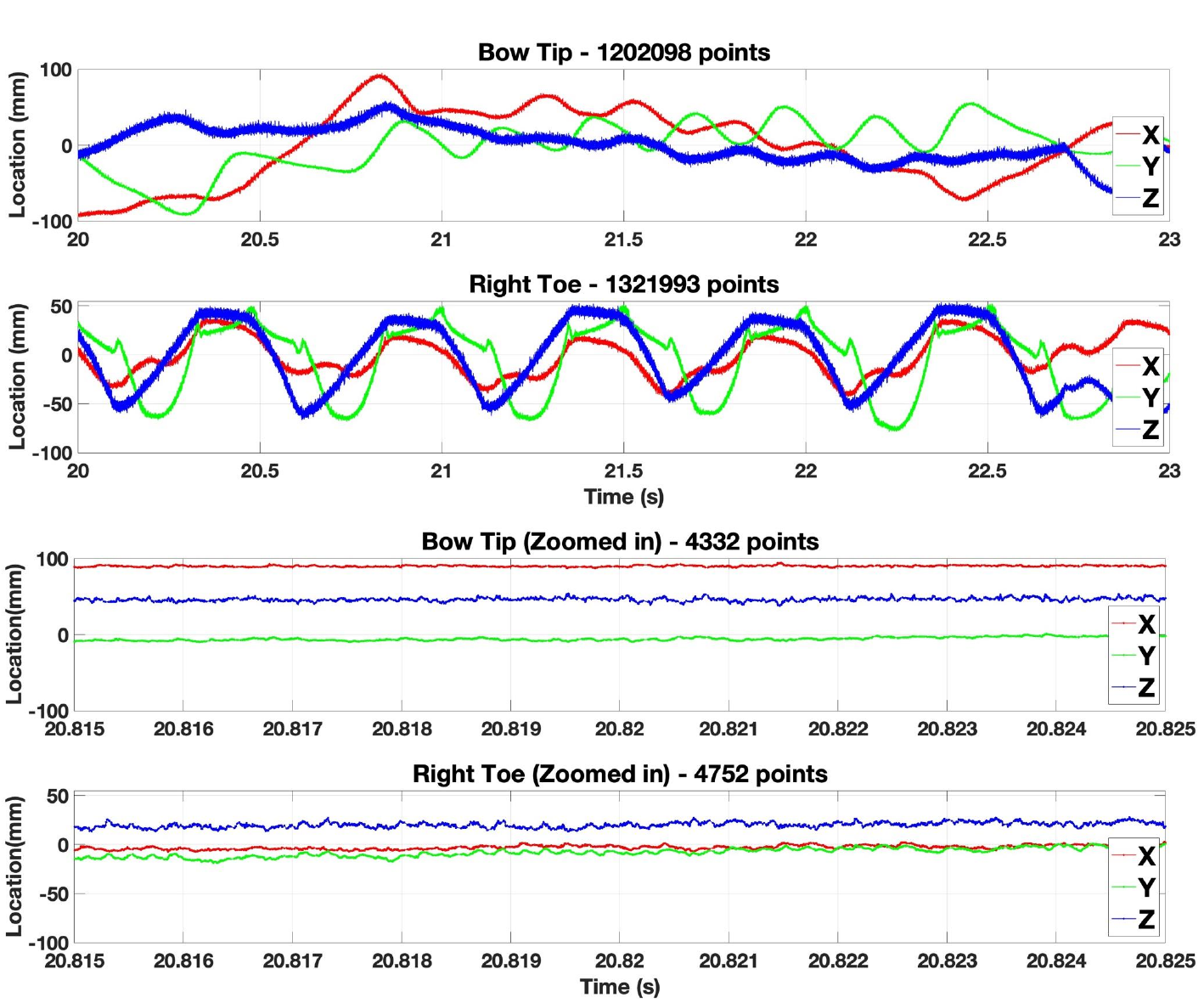}
	\end{tabular}
	\caption{
		\textbf{High-resolution 3D trajectories of the bow tip and right toe beacons.}
		For each beacon, the reconstructed $X$ (red), $Y$ (green), and $Z$ (blue) positions are plotted relative to their mean.
		The top two plots show the trace of position over a $3$~s trajectory, while the bottom plots are a 300$\times$ zoom in the temporal axis, showing 10~ms.
		This plot shows data at the temporal resolution of the underlying AEB tracker, $\sim$1~MHz.
		}\label{fig:3D_motion}
\end{figure}

\subsection{Motion Capture of a Violin Performance}

To demonstrate the performance of the proposed visual motion capture in a more realistic scenario, we perform 3D optical tracking of a violin (or `fiddle' in the folk music community) performance, using a pair of stereo event cameras to track nine smart beacons attached to the performer and instrument.
Four beacons are mounted on the musician: two on the knees and two on the feet.
Three beacons are mounted on the violin to recover its full 3D pose, and two beacons are mounted at either end of the bow, sufficient to locate its main axis.
The instrument beacons are attached to batons mounted on the fiddle to avoid contact with the instrument body and to enhance visibility from both cameras.
Figure~\ref{fig:3d-setup} (left) shows the fiddle, bow, and boots equipped with beacons, while (right) provides an overview of the stereo event cameras and the reference RGB camera used in the experiment.

In the demonstration, the AEB tracker reliably tracks all nine beacons.
Example 3D reconstruction frames are shown in Figure~\ref{fig:3d-demo}, where each beacon is drawn as a coloured dot, alongside the corresponding RGB camera reference frames at $t = 5.4$, $8.2$ and $37.2$~s.
Please refer to the supplementary video for a better visualisation of the full performance.

Figure~\ref{fig:3D_motion} illustrates the spatio-temporal resolution capability of the systems.
The bow and the right toe carry the fastest moving beacons in the performance, with the bow tip reaching $\sim$0.5~kpx/s and the right toe exceeding 0.8~kpx/s. The impulsive motion of the boot toe, which stops abruptly as it taps the board, leads to very fast changes in velocity.
A zoomed-in $10$~ms window of the same trajectories shows the high temporal resolution of the reconstruction.

\section{Conclusion}

In this paper, we present ATOC, an Asynchronous Tracking and Optical Communication system that combines modulated smart LED beacons with event-based vision for robust, low-latency optical communication, tracking and localisation.
Each beacon transmits data, transforming a conventional visual marker into a compact interface for sensing, communication, and temporal coordination.
The system is benchmarked at a reliable bit rate of 11.11~kb/s while tracking a beacon moving at up to 9.2~kpx/s in laboratory conditions.
In real-world conditions, transmission rates were reduced to 5~kb/s due to background noise events in order to guarantee 100\% data transmission.

Our indoor and outdoor results demonstrate that event cameras can provide more than high-speed visual sensing: they can support a shared channel for identification, communication, localisation, and synchronisation.
ATOC therefore offers a scalable, low-cost, and adaptable foundation for cooperative robotics, distributed sensing, motion capture, and smart infrastructure, where agents can exchange information simply by observing one another.
Moreover, its local and computationally lightweight processing architecture creates a path toward hardware-level implementation in emerging event sensors with pixel-level memory and computation, enabling increasingly compact and responsive visual communication systems.

\appendices
\section{Smart Beacons}\label{appendix: hardware}
The Adafruit FLORA (shown in Figure~\ref{fig:circuit and camera}(b), right) was initially used as an off-the-shelf prototyping platform because of its low cost and accessibility.
Based on the ATmega32U4 microcontroller, the device costs approximately US\$14.95.
However, experimental evaluation showed that its processing and switching capabilities restrict communication to relatively low transmission frequencies, making it unsuitable for the high-rate optical communication required in this work.

\begin{figure}[h!]
	\centering
	\begin{tabular}{c}
		\\	\includegraphics[width=0.98\linewidth]{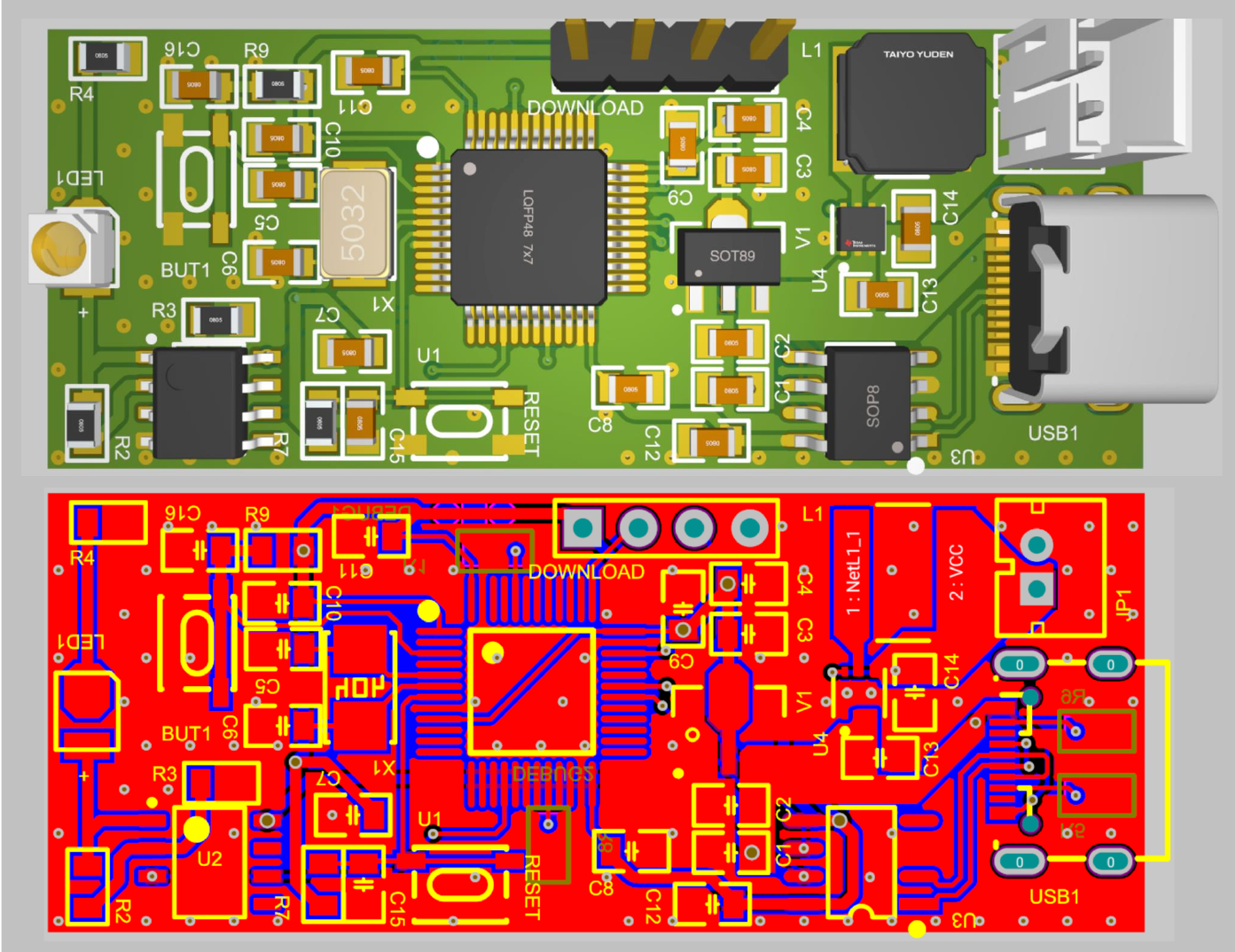}
	\end{tabular}
	\caption{
		Custom smart beacon PCB. Top: 3D rendering of the assembled board; bottom: PCB layout.
	}\label{fig:pcb}
\end{figure}

To overcome these limitations, we designed a custom smart beacon based on the STM32F103 microcontroller.
The total cost of the proposed hardware is approximately US\$25, including the battery.
It has a compact, lightweight form factor approximately half the size of a credit card.
The hardware design is intentionally simple, making it suitable for large-scale deployment.
The 3D rendering and PCB layout of the proposed beacon are shown in Figure~\ref{fig:pcb}.

\clearpage
\bibliographystyle{IEEEtran}
\bibliography{References}

@STRING{acra    = "Australasian Conf. Robot. Autom. (ACRA)" }

@STRING{arxiv   = "ar{X}iv e-prints" }

@STRING{case    = "{IEEE} Int. Conf. Automation Science and Engineering (CASE)" }

@STRING{computer= "{IEEE} Computer" }

@STRING{icra    = "{IEEE} Int. Conf. Robot. Autom. (ICRA)" }

@STRING{icuas   = "{IEEE} Int. Conf. Unmanned Aircraft Syst. (ICUAS)" }

@STRING{ieee    = "Proc. {IEEE}" }

@STRING{iros    = "IEEE/RSJ Int. Conf. Intell. Robot. Syst. (IROS)" }

@STRING{science = "Science" }

@Article{Benosman14tnnls,
	author        = {Ryad Benosman and Charles Clercq and Xavier Lagorce and
	Sio-Hoi Ieng and Chiara Bartolozzi},
	title         = {Event-Based Visual Flow},
	journal       = {IEEE Transactions on Neural Networks and Learning Systems},
	year          = 2014,
	volume        = 25,
	number        = 2,
	pages         = {407--417},
	doi           = {10.1109/TNNLS.2013.2273537}
}

@Book{Hartley03book,
	author        = {Richard Hartley and Andrew Zisserman},
	title         = {Multiple View Geometry in Computer Vision},
	publisher     = {Cambridge University Press},
	year          = 2003,
	note          = {2nd Edition},
	doi           = {10.1017/CBO9780511811685}
}

@InProceedings{Wang19acra,
	author        = {Ziwei Wang and Yonhon Ng and Pieter van Goor and Robert
	Mahony},
	title         = {Event Camera Calibration of Per-pixel Biased Contrast
	Threshold},
	booktitle     = {Australasian Conference of Robotics and Automation (ACRA)},
	year          = 2019
}

@inproceedings{scheerlinck2018continuous,
	title={Continuous-time intensity estimation using event cameras},
	author={Scheerlinck, Cedric and Barnes, Nick and Mahony, Robert},
	booktitle={Asian Conference on Computer Vision},
	pages={308--324},
	year={2018},
	organization={Springer}
}

@inproceedings{imai2016performance,
	title={Performance evaluation of high-speed visible light communication combining low-speed image sensor and polygon mirror in an outdoor environment},
	author={Imai, Yoshihito and Ebihara, Tadashi and Mizutani, Koichi and Wakatsuki, Naoto},
	booktitle={2016 Eighth International Conference on Ubiquitous and Future Networks (ICUFN)},
	pages={51--55},
	year={2016},
	organization={IEEE}
}

@inproceedings{censi2013low,
	title={Low-latency localization by active LED markers tracking using a dynamic vision sensor},
	author={Censi, Andrea and Strubel, Jonas and Brandli, Christian and Delbruck, Tobi and Scaramuzza, Davide},
	booktitle={2013 IEEE/RSJ International Conference on Intelligent Robots and Systems},
	pages={891--898},
	year={2013},
	organization={IEEE}
}

@inproceedings{faessler2014monocular,
	title={A monocular pose estimation system based on infrared leds},
	author={Faessler, Matthias and Mueggler, Elias and Schwabe, Karl and Scaramuzza, Davide},
	booktitle={2014 IEEE international conference on robotics and automation (ICRA)},
	pages={907--913},
	year={2014},
	organization={IEEE}
}

@inproceedings{walter2018fast,
	title={Fast mutual relative localization of uavs using ultraviolet led markers},
	author={Walter, Viktor and Saska, Martin and Franchi, Antonio},
	booktitle={2018 International Conference on Unmanned Aircraft Systems (ICUAS)},
	pages={1217--1226},
	year={2018},
	organization={IEEE}
}

@inproceedings{walter2018mutual,
	title={Mutual localization of uavs based on blinking ultraviolet markers and 3d time-position hough transform},
	author={Walter, Viktor and Staub, Nicolas and Saska, Martin and Franchi, Antonio},
	booktitle={2018 IEEE 14th International Conference on Automation Science and Engineering (CASE)},
	pages={298--303},
	year={2018},
	organization={IEEE}
}

@article{bosch2016close,
	title={Close-range tracking of underwater vehicles using light beacons},
	author={Bosch, Josep and Gracias, Nuno and Ridao, Pere and Isteni{\v{c}}, Klemen and Ribas, David},
	journal={Sensors},
	volume={16},
	number={4},
	pages={429},
	year={2016},
	publisher={Multidisciplinary Digital Publishing Institute}
}

@inproceedings{dias2016board,
	title={On-board vision-based 3D relative localization system for multiple quadrotors},
	author={Dias, Duarte and Ventura, Rodrigo and Lima, Pedro and Martinoli, Alcherio},
	booktitle={2016 IEEE International Conference on Robotics and Automation (ICRA)},
	pages={1181--1187},
	year={2016},
	organization={Ieee}
}

@article{gracias2015pose,
	title={Pose estimation for underwater vehicles using light beacons},
	author={Gracias, Nuno and Bosch, Josep and Karim, Mohammad Ehsanul},
	journal={IFAC-PapersOnLine},
	volume={48},
	number={2},
	pages={70--75},
	year={2015},
	publisher={Elsevier}
}

@article{chen2020novel,
	title={A novel visible light positioning system with event-based neuromorphic vision sensor},
	author={Chen, Guang and Chen, Wenkai and Yang, Qianyi and Xu, Zhongcong and Yang, Longyu and Conradt, J{\"o}rg and Knoll, Alois},
	journal={IEEE Sensors Journal},
	volume={20},
	number={17},
	pages={10211--10219},
	year={2020},
	publisher={IEEE}
}

@article{joubert2019characterization,
	title={Characterization setup for event-based imagers applied to modulated light signal detection},
	author={Joubert, Damien and H{\'e}bert, Mathieu and Konik, Hubert and Lavergne, Christophe},
	journal={Applied optics},
	volume={58},
	number={6},
	pages={1305--1317},
	year={2019},
	publisher={Optical Society of America}
}

@inproceedings{perez2019optical,
	title={Optical Wireless Camera Communications using Neuromorphic Vision Sensors},
	author={Perez-Ramirez, Javier and Roberts, Richard D and Navik, Ankit P and Muralidharan, Nagarajan and Moustafa, Hassnaa},
	booktitle={2019 IEEE International Conference on Communications Workshops (ICC Workshops)},
	pages={1--6},
	year={2019},
	organization={IEEE}
}

@article{saeed2019optical,
	title={Optical camera communications: Survey, use cases, challenges, and future trends},
	author={Saeed, Nasir and Guo, Shuaishuai and Park, Ki-Hong and Al-Naffouri, Tareq Y and Alouini, Mohamed-Slim},
	journal={Physical Communication},
	volume={37},
	pages={100900},
	year={2019},
	publisher={Elsevier}
}

@inproceedings{nguyen2016high,
	title={High-rate flicker-free screen-camera communication with spatially adaptive embedding},
	author={Nguyen, Viet and Tang, Yaqin and Ashok, Ashwin and Gruteser, Marco and Dana, Kristin and Hu, Wenjun and Wengrowski, Eric and Mandayam, Narayan},
	booktitle={IEEE INFOCOM 2016-The 35th Annual IEEE International Conference on Computer Communications},
	pages={1--9},
	year={2016},
	organization={IEEE}
}

@inproceedings{rajagopal2014visual,
	title={Visual light landmarks for mobile devices},
	author={Rajagopal, Niranjini and Lazik, Patrick and Rowe, Anthony},
	booktitle={IPSN-14 proceedings of the 13th international symposium on information processing in sensor networks},
	pages={249--260},
	year={2014},
	organization={IEEE}
}

@inproceedings{lee2015rollinglight,
	title={Rollinglight: Enabling line-of-sight light-to-camera communications},
	author={Lee, Hui-Yu and Lin, Hao-Min and Wei, Yu-Lin and Wu, Hsin-I and Tsai, Hsin-Mu and Lin, Kate Ching-Ju},
	booktitle={Proceedings of the 13th annual international conference on mobile systems, applications, and services},
	pages={167--180},
	year={2015}
}

@inproceedings{roberts2013undersampled,
	title={Undersampled frequency shift ON-OFF keying (UFSOOK) for camera communications (CamCom)},
	author={Roberts, Richard D},
	booktitle={2013 22nd Wireless and Optical Communication Conference},
	pages={645--648},
	year={2013},
	organization={IEEE}
}

@inproceedings{aoyama2015visible,
	title={Visible light communication using a conventional image sensor},
	author={Aoyama, Hideki and Oshima, Mitsuaki},
	booktitle={2015 12th Annual IEEE Consumer Communications and Networking Conference (CCNC)},
	pages={103--108},
	year={2015},
	organization={IEEE}
}

@inproceedings{aoyama2015line,
	title={Line scan sampling for visible light communication: Theory and practice},
	author={Aoyama, Hideki and Oshima, Mitsuaki},
	booktitle={2015 IEEE International Conference on Communications (ICC)},
	pages={5060--5065},
	year={2015},
	organization={IEEE}
}

@article{nguyen2016region,
	title={Region-of-interest signaling vehicular system using optical camera communications},
	author={Nguyen, Trang and Islam, Amirul and Jang, Yeong Min},
	journal={IEEE Photonics Journal},
	volume={9},
	number={1},
	pages={1--20},
	year={2016},
	publisher={IEEE}
}

@inproceedings{luo2014undersampled,
	title={Undersampled phase shift ON-OFF keying for camera communication},
	author={Luo, Pengfei and Ghassemlooy, Zabih and Le Minh, Hoa and Tang, Xuan and Tsai, Hsin-Mu},
	booktitle={2014 Sixth International Conference on Wireless Communications and Signal Processing (WCSP)},
	pages={1--6},
	year={2014},
	organization={IEEE}
}

@inproceedings{roberts2014automotive,
	title={Automotive comphotogrammetry},
	author={Roberts, Richard D},
	booktitle={2014 IEEE 79th vehicular technology conference (VTC Spring)},
	pages={1--5},
	year={2014},
	organization={IEEE}
}

@article{teli2018high,
	title={High-speed optical camera V2V communications using selective capture},
	author={Teli, Shivani and Cahyadi, Willy Anugrah and Chung, Yeon Ho},
	journal={Photonic network communications},
	volume={36},
	number={2},
	pages={210--216},
	year={2018},
	publisher={Springer}
}

@article{ebihara2015layered,
	title={Layered transmission of space-time coded signals for image-sensor-based visible light communications},
	author={Ebihara, Katsunori and Kamakura, Koji and Yamazato, Takaya},
	journal={Journal of lightwave technology},
	volume={33},
	number={20},
	pages={4193--4206},
	year={2015},
	publisher={IEEE}
}

@inproceedings{wang2021asynchronous,
	title={An asynchronous kalman filter for hybrid event cameras},
	author={Wang, Ziwei and Ng, Yonhon and Scheerlinck, Cedric and Mahony, Robert},
	booktitle={Proceedings of the IEEE/CVF International Conference on Computer Vision},
	pages={448--457},
	year={2021}
}

@book{shakespeare1989william,
  title={William Shakespeare: the complete works},
  Note={Sonnet 18, ``Shall I compare thee to a summer’s day''},
  author={Shakespeare, William and others},
  year={1989},
  publisher={Barnes \& Noble Publishing}
}

@inproceedings{oike2003smart,
	title={A smart image sensor with high-speed feeble ID-beacon detection for augmented reality system},
	author={Oike, Yusuke and Ikeda, Makoto and Asada, Kunihiro},
	booktitle={ESSCIRC 2004-29th European Solid-State Circuits Conference (IEEE Cat. No. 03EX705)},
	pages={125--128},
	year={2003},
	organization={IEEE}
}

@InProceedings{wang22icra,
	author        = {Ziwei Wang and Dingran Yuan and Yonhon Ng and Robert Mahony},
	title         = {A Linear Comb Filter for Event Flicker Removal},
	booktitle     = {"International Conference on Robotics and Automation (ICRA)" },
	year          = {2022}
}

@inproceedings{wang2021stereo,
	title={Stereo Hybrid Event-Frame (SHEF) Cameras for 3D Perception},
	author={Wang, Ziwei and Pan, Liyuan and Ng, Yonhon and Zhuang, Zheyu and Mahony, Robert},
	booktitle={2021 IEEE/RSJ International Conference on Intelligent Robots and Systems (IROS)},
	year={2021},
	organization={IEEE}
}

@inproceedings{wang2022smart,
	title={Smart Visual Beacons with Asynchronous Optical Communications using Event Cameras},
	author={Wang, Ziwei and Ng, Yonhon and Henderson, Jack and Mahony, Robert},
	booktitle={2022 IEEE/RSJ International Conference on Intelligent Robots and Systems (IROS)},
	pages={3793--3799},
	year={2022},
	organization={IEEE}
}

@article{wang2024aeb,
	title={Asynchronous Blob Tracker for Event Cameras},
	author={Wang, Ziwei and Molloy, Timothy and van Goor, Pieter and Mahony, Robert},
	journal={IEEE Transactions on Robotics},
	year={2024},
	publisher={IEEE}
}

@ARTICLE{Hamming_1950,
	author={Hamming, R. W.},
	journal={The Bell System Technical Journal}, 
	title={Error detecting and error correcting codes}, 
	year={1950},
	volume={29},
	number={2},
	pages={147-160},
	doi={10.1002/j.1538-7305.1950.tb00463.x}}

@INPROCEEDINGS{Saiz_2016,
	
	author={Saiz-Adalid, Luis-J. and Gil, Pedro and Ruiz, Juan-Carlos and Gracia-Morán, Joaquín and Gil-Tomás, Daniel and Baraza-Calvo, J.-Carlos},
	
	booktitle={2016 12th European Dependable Computing Conference (EDCC)}, 
	
	title={Ultrafast Error Correction Codes for Double Error Detection/Correction}, 
	
	year={2016},
	
	volume={},
	
	number={},
	
	pages={108-119},
	
	doi={10.1109/EDCC.2016.28}}

@article{immink1990runlength,
	title={Runlength-limited sequences},
	author={Immink, Kees A Schouhamer},
	journal={Proceedings of the IEEE},
	volume={78},
	number={11},
	pages={1745--1759},
	year={1990},
	publisher={IEEE}
}

@article{apps2025aemot,
  title={Asynchronous Multi-Object Tracking with an Event Camera},
  author={Apps, Angus and Wang, Ziwei and Perejogin, Vladimir and Molloy, Timothy and Mahony, Robert},
  journal={arXiv preprint arXiv:2505.08126},
  year={2025}
}

@article{aranda2024enhancing,
	title={Enhancing Computational Efficiency in Event-Based Optical Camera Communication Using N-Pulse Modulation},
	author={Aranda, Jaime and Guerra, Victor and Rabadan, Jose and Perez-Jimenez, Rafael},
	journal={Electronics},
	volume={13},
	number={6},
	pages={1047},
	year={2024},
	publisher={MDPI}
}

@article{von2024dynamic,
	title={Dynamic event-based optical identification and communication},
	author={von Arnim, Axel and Lecomte, Jules and Borras, Naima Elosegui and Wo{\'z}niak, Stanis{\l}aw and Pantazi, Angeliki},
	journal={Frontiers in Neurorobotics},
	volume={18},
	pages={1290965},
	year={2024},
	publisher={Frontiers Media SA}
}

@inproceedings{wang2024towards,
	title={Towards High-Speed Passive Visible Light Communication with Event Cameras and Digital Micro-Mirrors},
	author={Wang, Yanxiang and Shen, Yiran and Xu, Kenuo and Hassan, Mahbub and Zhao, Guangrong and Xu, Chenren and Hu, Wen},
	booktitle={Proceedings of the 22nd ACM Conference on Embedded Networked Sensor Systems},
	pages={704--717},
	year={2024}
}

@article{su2024motion,
	title={Motion-Aware Optical Camera Communication with Event Cameras},
	author={Su, Hang and Gao, Ling and Liu, Tao and Kneip, Laurent},
	journal={IEEE Robotics and Automation Letters},
	year={2024},
	publisher={IEEE}
}

@inproceedings{kitade2024bicode,
	title={Bicode: A Hybrid Blinking Marker System for Event Cameras},
	author={Kitade, Takuya and Yamada, Wataru and Ochiai, Keiichi and Imai, Michita},
	booktitle={2024 IEEE International Conference on Robotics and Automation (ICRA)},
	pages={2939--2945},
	year={2024},
	organization={IEEE}
}

@inproceedings{ebmer2024real,
	title={Real-time 6-dof pose estimation by an event-based camera using active led markers},
	author={Ebmer, Gerald and Loch, Adam and Vu, Minh Nhat and Mecca, Roberto and Haessig, Germain and Hartl-Nesic, Christian and Vincze, Markus and Kugi, Andreas},
	booktitle={Proceedings of the IEEE/CVF Winter Conference on Applications of Computer Vision},
	pages={8137--8146},
	year={2024}
}

@article{bauersfeld2025monocular,
	title={A Monocular Event-Camera Motion Capture System},
	author={Bauersfeld, Leonard and Scaramuzza, Davide},
	journal={arXiv preprint arXiv:2502.12113},
	year={2025}
}

@Misc{2025_AdaFruit,
  author = {{AdaFruit Industries}},
  note   = {visited 27 September 2025},
  title  = {FLORA - Wearable electronic platform: Arduino-compatible - v3},
  year   = {2025},
  url    = {https://www.adafruit.com/product/659?srsltid=AfmBOorrh2kPGVsaE1n1knCsigBKepdgcvVfZgh0ZWGmcFt47-CNo3YM},
}

@article{luo2018undersampled,
  title={Undersampled-based modulation schemes for optical camera communications},
  author={Luo, Pengfei and Zhang, Min and Ghassemlooy, Zabih and Zvanovec, Stanislav and Feng, Shulan and Zhang, Philipp},
  journal={IEEE Communications Magazine},
  volume={56},
  number={2},
  pages={204--212},
  year={2018},
  publisher={IEEE}
}

@article{cahyadi2020optical,
  title={Optical camera communications: Principles, modulations, potential and challenges},
  author={Cahyadi, Willy Anugrah and Chung, Yeon Ho and Ghassemlooy, Zabih and Hassan, Navid Bani},
  journal={Electronics},
  volume={9},
  number={9},
  pages={1339},
  year={2020},
  publisher={MDPI}
}

@inproceedings{tofighi2026fast,
  title={Fast and Robust Event-based Optical Communication for Aerial Robots via Active LED Markers},
  author={Tofighi, Nafiseh Jabbari and Robic, Maxime and Caracotte, Jordan and Vasseur, Pascal and Morbidi, Fabio},
  booktitle={2026 International Conference on Unmanned Aircraft Systems (ICUAS)},
  pages={1314--1321},
  year={2026},
  organization={IEEE}
}

@inproceedings{soga2025evaluation,
  title={Evaluation of mobile environment for vehicular visible light communication using multiple leds and event cameras},
  author={Soga, Ryota and Shiba, Shintaro and Kong, Quan and Kobori, Norimasa and Shimizu, Tsukasa and Lu, Shan and Yamazato, Takaya},
  booktitle={2025 IEEE Intelligent Vehicles Symposium (IV)},
  pages={964--969},
  year={2025},
  organization={IEEE}
}

@article{soga2026real,
  title={Real-world Latency Analysis of Vehicular Visible Light Communication with Multiple LED Transmitters and an Event-Based Camera},
  author={Soga, Ryota and Shimizu, Tsukasa and Shiba, Shintaro and Kong, Quan and Lu, Shan and Yamazato, Takaya},
  journal={arXiv preprint arXiv:2605.05541},
  year={2026}
}

@misc{CAD_model_violin,
  author = {UrielAngel},
  howpublished = {\url{https://www.thingiverse.com/thing:1845197/files}},
  title = {Violin},
  year = {2016},
  note = {Accessed: 2026-03-23}
}

@misc{CAD_model_bow,
  author = {Eric Slotty},
  howpublished = {\url{https://grabcad.com/library/relaxed-cello-bow-1}},
  title = {Relaxed cello bow},
  year = {2023},
  note = {Accessed: 2026-03-23}
}

@misc{CAD_model_boot,
  author = {ObiWenHopibi},
  howpublished = {\url{https://www.thingiverse.com/thing:4785094/files}},
  title = {Work boot},
  year = {2021},
  note = {Accessed: 2026-03-23}
}

\begin{IEEEbiography}[{\includegraphics[width=1in,height=1.25in,clip,keepaspectratio]{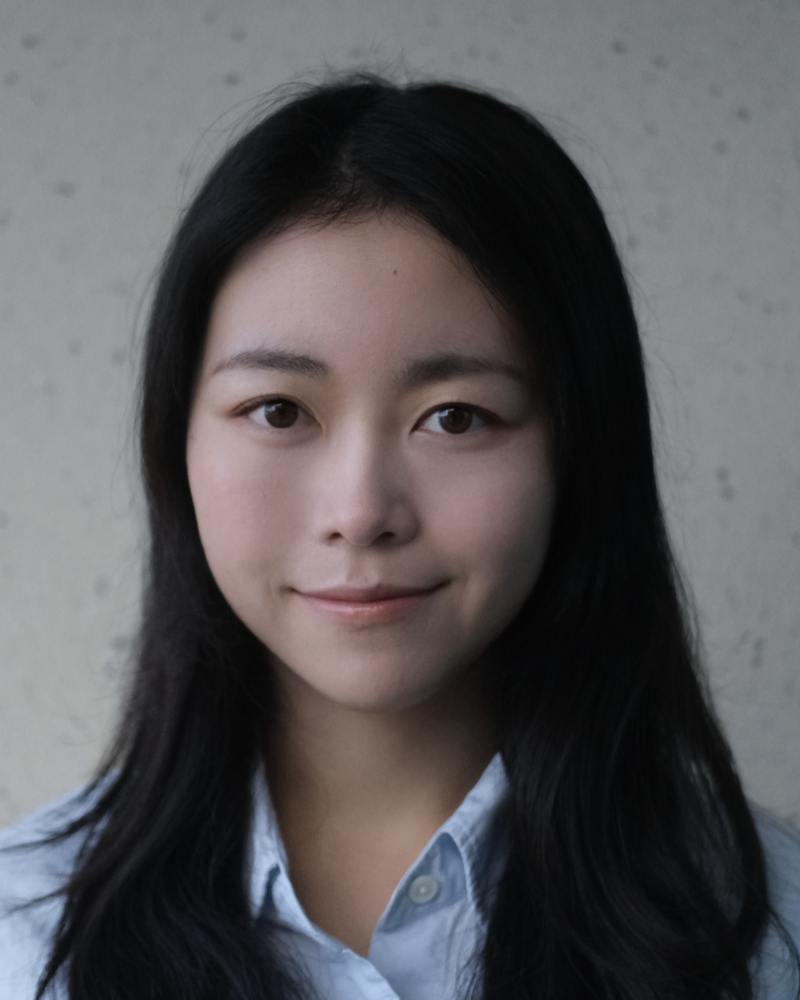}}]{Ziwei Wang}
	(Student Member, IEEE) received the B.E. degree in mechatronics from the Australian National University (ANU), Canberra, Australia, in 2019, and the Ph.D. degree from the Systems Theory and Robotics (STR) group at the School of Engineering, ANU, in 2025.
	Her research interests include event-based vision, asynchronous image processing, autonomous driving, and embodied AI, across a wide range of robotic applications.
\end{IEEEbiography}

\begin{IEEEbiography}[{\includegraphics[width=1in,height=1.25in,clip,keepaspectratio]{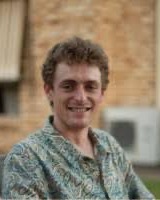}}]{Angus Apps}
	is a Ph.D. student in the Systems Theory and Robotics (STR) group at the School of Engineering, Australian National University (ANU), Canberra, Australia. His research interests include event-based vision and asynchronous image processing for robotics applications.
\end{IEEEbiography}

\begin{IEEEbiography}[{\includegraphics[width=1in,height=1.25in,clip,keepaspectratio]{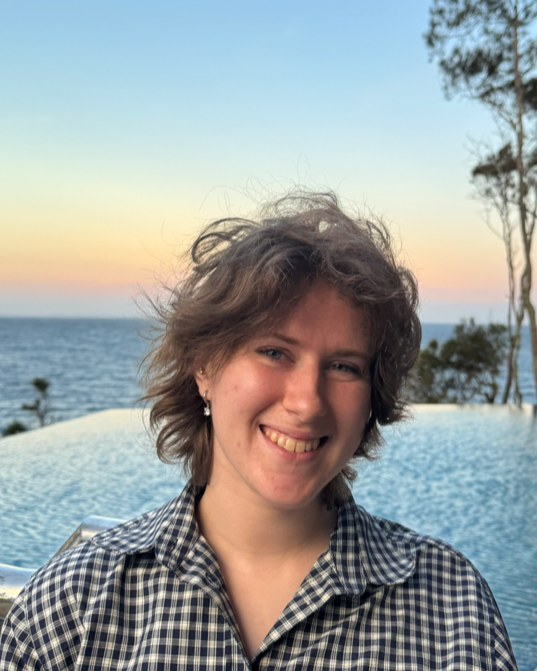}}]{Holly Battisson}
	received the B.E. (Honours) degree from the Australian National University (ANU), Canberra, Australia, in 2024. She completed her honours project in the Systems Theory and Robotics (STR) group at the School of Engineering, ANU. Her research interests include event cameras and computer vision.
\end{IEEEbiography}

\begin{IEEEbiography}[{\includegraphics[width=1in,height=1.25in,clip,keepaspectratio]{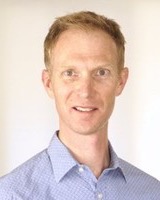}}]{Iain Guilliard}
	is an Adjunct Fellow in the Systems Theory and Robotics (STR) group at the School of Engineering, Australian National University (ANU), Canberra, Australia. His research interests include artificial intelligence, planning under uncertainty, and optimisation, with applications to autonomous soaring for unmanned aerial vehicles and to intelligent transport systems.
\end{IEEEbiography}

\begin{IEEEbiography}[{\includegraphics[width=1in,height=1.25in,clip,keepaspectratio]{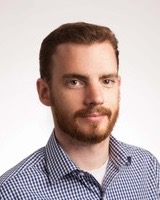}}]{Timothy L. Molloy} (Member, IEEE) was born in Emerald, Australia. He received the BE and PhD degrees from the Queensland University of Technology (QUT) in 2010 and 2015, respectively. He is currently a Senior Lecturer in the School of Engineering at the Australian National University (ANU). Prior to joining ANU, he was an Advance Queensland Research Fellow at QUT (2017-2019), and a Research Fellow at the University of Melbourne (2020-2022). His interests include signal processing and information theory for robotics and control.
	Dr. Molloy is the recipient of a QUT University Medal, a QUT Outstanding Doctoral Thesis Award, a 2018 Boeing Wirraway Team Award, and an Advance Queensland Early-Career Fellowship.\end{IEEEbiography}

\begin{IEEEbiography}[{\includegraphics[width=1in,height=1.25in,clip,keepaspectratio]{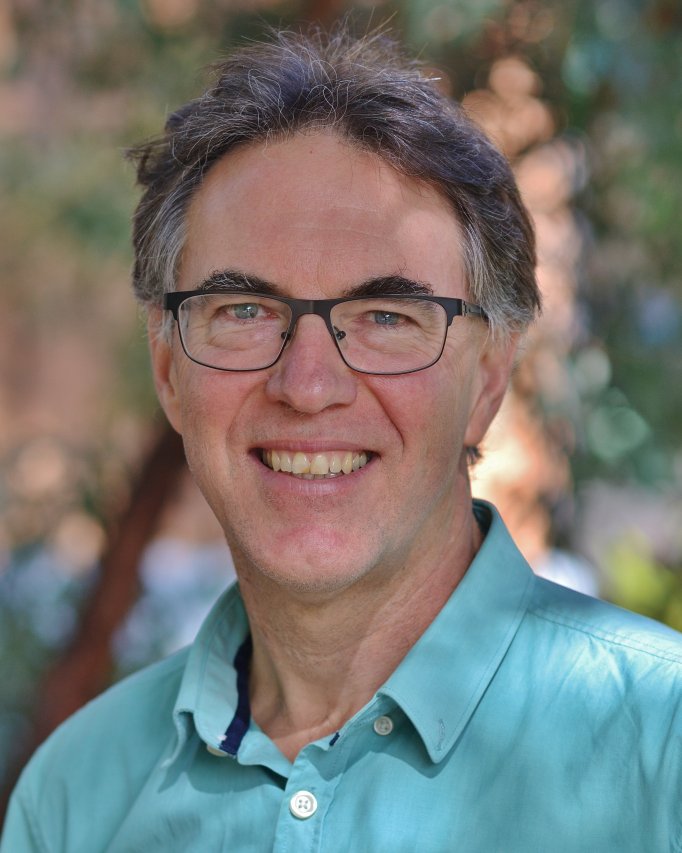}}]{Robert Mahony} (Fellow, IEEE) is a Professor in the School of Engineering at the Australian National University. He is the lead of the Systems Theory and Robotics (STR) group. He received his BSc in 1989 (applied mathematics and geology) and his PhD in 1995 (systems engineering) both from the Australian National University. He is a fellow of the IEEE. His research interests are in non-linear systems theory with applications in robotics and computer vision. He is known for his work in aerial robotics, geometric observer design, robotic vision, and optimisation on matrix manifolds.
\end{IEEEbiography}

\end{document}